\documentclass[twoside,11pt]{article}
\usepackage{graphicx, wrapfig, setspace, booktabs}
\usepackage{subfigure}
\usepackage{amssymb}
\usepackage{amsmath}
\usepackage{amsthm}
\usepackage{caption}
\usepackage{calc, accents}
\usepackage{comment}
\usepackage{xcolor}
\usepackage{arydshln}

\newcommand{\bx}{\mathbf{x}}
\newcommand{\by}{\mathbf{y}}

\newcommand{\bA}{{\mathbf A}}

\newcommand{\bI}{{\mathbf I}}

\newcommand{\bM}{{\mathbf M}}
\newcommand{\bQ}{{\mathbf Q}}
\newcommand{\bP}{{\mathbf P}}

\newcommand{\bX}{{\mathbf X}}

\newcommand{\bZ}{{\mathbf Z}}

\newcommand{\bfp}{{\mathbf p}}
\newcommand{\bfq}{{\mathbf q}}

\newcommand{\br}{{\mathbf r}}

\newcommand{\bs}{{\mathbf s}}

\newcommand{\bTheta} {\boldsymbol{\Theta}}

\newcommand{\dbtilde}[1]{\accentset{\approx}{#1}}

\newtheorem{prop}{Proposition} 
\newtheorem{assu}{Assumption}
\newtheorem{theo}{Theorem}
\newtheorem{example}{Example} 
\newtheorem{lem}[example]{Lemma}

\def\be{\begin{equation}}
\def\ee{\end{equation}}
\def\bea{\begin{eqnarray}}
\def\eea{\end{eqnarray}}
\def\T{{ \mathrm{\scriptscriptstyle T} }}
\def\nn{\nonumber}

\newcommand{\bftab}{\fontseries{b}\selectfont}

\usepackage{jmlr2e}

\begin{document}

\title{Radial Neighborhood Smoothing Recommender System}

\author{\name Zerui Zhang \email zeruiz@iastate.edu \\
       \addr Department of Statistics and\\
       Bioinformatics and Computational Biology Program \\
       Iowa State University\\
       Ames, IA 50010, USA
       \AND
       \name Yumou Qiu \email qiuyumou@math.pku.edu.cn \\
       \addr School of Mathematical Sciences and \\
       Center for Statistical Science\\
       Peking University\\
       Beijing, 100871, China}

\maketitle

\begin{abstract}
Recommender systems inherently exhibit a low-rank structure in latent space. A key challenge is to define meaningful and measurable distances in the latent space to capture user-user, item-item, user-item relationships effectively. In this work, we establish that distances in the latent space can be systematically approximated using row-wise and column-wise distances in the observed matrix, providing a novel perspective on distance estimation. To refine the distance estimation, we introduce the correction based on empirical variance estimator to account for noise-induced non-centrality. The novel distance estimation enables a more structured approach to constructing neighborhoods, leading to the Radial Neighborhood Estimator (RNE), which constructs neighborhoods by including both overlapped and partially overlapped user-item pairs and employs neighborhood smoothing via localized kernel regression to improve imputation accuracy. We provide the theoretical asymptotic analysis for the proposed estimator. We perform evaluations on both simulated and real-world datasets, demonstrating that RNE achieves superior performance compared to existing collaborative filtering and matrix factorization methods. While our primary focus is on distance estimation in latent space, we find that RNE also mitigates the ``cold-start'' problem.
\end{abstract}


\section{Introduction}

Matrix completion, widely known in the industry as recommendation systems, is of great interest in a variety of fast-growing fields in e-commerce, medicine \citep{ni-24-time-series}, finance \citep{Yang2024a} and social networks \citep{ding2025trade}: consider the measured data can form a sparse matrix, i.e., only a small fraction of the data can be observed, and the task is to predict the unobserved entries and thus filling the whole matrix \citep{rennie2005fast, candes2009exact, li2025encoder, li2025finecir}. Some prevalent examples of recommender systems include \cite{harper2015movielens, Bertin-Mahieux2011, linden2003amazon}, etc.

Two primary streams have been established in earlier works to tackle the recommender system training: matrix factorization based approaches and neighborhood based approaches. Matrix factorization admits that all matrices can be uniquely factorized by singular value decomposition such that the row and column singular vectors can be recovered accurately from the partially observed, noisy matrix and subsequently estimate the true preference matrix. Inspired by the nuclear norm criterion \citep{cai2010singular,candes2009exact,candes2010power}, singular-value-thresholding algorithm \citep{cai2010singular}, maximum margin matrix factorization \citep{rennie2005fast}, soft-impute \citep{mazumder2010spectral}, among a line of ongoing work have been proposed and widely used. For neighborhood based approaches, collaborative filtering \citep{goldberg1992using} is featured for its simplicity and ability to scale. Item-item paradigm and user-user paradigm recommend items according to the exploitation of pairwise information in a pool of similar items or users, which have been recognized as ``neighbors'' \citep{ning2015comprehensive}. There have been improvements over the basic algorithm such as combining both users and items data as weight interpolations for final prediction \citep{wang2006unifying,bell2007scalable}, and retrieving user-item interactions by Taylor expansion \citep{song2016blind} or smoothing structure with additional external covariates \citep{dai2019smooth}. Recent research leverages the Large Language models (LLM) to explore the user intent latent structure to predict the next item of interest through the LSTM-generated prompts \citep{xu2025enhancinguserintentrecommendation, zhong2025narrative}.

Several challenges arise in recommender systems. When applying neighborhood-based approaches, one major limitation is that the number of effective neighbors is often restricted, which can hinder similarity-based predictions. Correlation-based similarity measures require multiple overlapping ratings between users or items \citep{goldberg1992using, balabanovic1997fab, billsus2000user, song2016blind}, while other approaches rely on external variables to define distances \citep{yu2022nonparametric}. These methods struggle in scenarios where a new user or item lacks prior interactions, leading to the cold-start problem. Moreover, existing models often fail to fully incorporate user-item interactions and hierarchical relationships. For instance, an individual’s friends typically share more similar preferences than randomly selected users, and even second-degree connections (friends of friends) may exhibit resemblance in tastes, despite not having rated the same items \citep{dai2019smooth, borgs2017thy}. Matrix factorization techniques, on the other hand, have their inherent drawbacks. One key limitation is that they capture only global latent features, overlooking localized similarity patterns among neighbors that could enhance prediction accuracy. Furthermore, predictions derived solely from latent features are often highly abstract and lack direct interpretability, making it challenging to provide meaningful explanations for recommendations.

We propose a novel method that integrates a latent variable model derived from singular value decomposition with neighbor pooling in a nonparametric smoothing framework. Our contributions are fourfold. First, we introduce a distance measure in latent space that approximates similarity using row- and column-wise $L_2$ norms based solely on observed ratings. This formulation bridges the gap between neighborhood-based methods and matrix factorization approaches, allowing our method to inherit the strengths of both. Second, we construct a radial neighborhood set, which captures both direct and indirect relationships among user-item pairs, thereby enriching local structure information. Third, our method relaxes the constraints on distance computation, eliminating requirements on the minimum number of overlapping observed ratings and the availability of additional covariates. The flexibility of the nonparametric framework allows it to effectively model network structures. Finally, while our primary focus is on distance measure in latent space and the neighbor set construction, we observe that our method also mitigates the ``cold-start'' problem by leveraging multi-level neighborhood to incorporate a great number of observed ratings. We validate our approach through both theoretical analysis and empirical evaluations, demonstrating its superior predictive performance and consistency.

The article is organized as follows. In Section~\ref{section2}, we will establish our model and distance estimation from the latent space to the observed matrix. We will also discuss the related methods. In Section~\ref{section3}, we will establish the theoretical properties of our proposed estimator. In Section~\ref{section4}, we will conduct simulation studies and several real examples to evaluate our proposed approach under different scenarios.


\section{Method}\label{section2}

\subsection{Latent factors by singular value decomposition}

We first provide the notations in a general recommender system problem. Consider a $n \times m$ matrix $\bA$ representing observed $n$ users' preference ratings on $m$ items, where $\bA_{j}$ and $\bA_{(j)}$ denote the $j$th column and $j$th row of $\bA$, i.e., the ratings received by item $j$ and the ratings provided by user $j$, correspondingly. Each entry of $\bA$ denoted by $a_{u, i}$ represents the specific ratings from the user-item pair $(u,i)$, and we can define the set for user-item pairs with observed ratings, $\Omega = \{(u,i):\exists~ a_{u,i}, u\in[n], i\in [m]\}$. Let $\bZ=(z_{u, i})_{n \times m}$ be the underlying scores by user-item pairs where $z_{u, i} \in [\eta_1,\eta_2]$ for all $u \in [n]$ and $i \in [m]$. Note that $\bZ$ is an unobserved yet complete matrix and $\bA$ is partially observed with a high missing percentage. We consider a simple model as
\bea
    a_{u,i}=z_{u,i}+\epsilon_{u,i}\quad \mathrm{for}\ u\in[n],\ i\in[m], \label{model}
\eea
where $\epsilon_{u,i}$ is independently and identically distributed random errors with zero mean and finite variance $\sigma^2$ for all $u\in[n]$ and $i\in[m]$. To elaborate on the construction of $\bZ$, we apply singular value decomposition as
\bea
    \bZ = \bP \bTheta \bQ^{\T} = \begin{bmatrix}\bfp_1 \cdots \bfp_k\end{bmatrix}\begin{bmatrix}
    c_{1} & & \\
    & \ddots & \\
    & & c_{k}
  \end{bmatrix}\begin{bmatrix}
    \bfq_{1}^{\T} \\
    \vdots \\
    \bfq_k^{\T}
  \end{bmatrix}, \label{model_svd}
\eea
where $\bP\in\mathbb{R}^{n\times k}$ is the left singular matrix such that $\bP^{\T}\bP=\bI_k$, which represents the relationship between users and latent factors. Let $\bfp_{(j)}$ denote the $j$th row of $\bP$. $\bTheta\in\mathbb{R}^{k\times k}$ is a  diagonal matrix, which describes the strength of each latent factor. $\bQ^{\T}\in\mathbb{R}^{k\times m}$ is the right singular matrix such that $\bQ^{\T}\bQ=\bI_k$, which indicates the relationship between items and latent factors. Let $\bfq_{j}$ denotes the $j$th column of $\bQ^{\T}$. We want to specify some general assumptions for $\bZ$ here:

\begin{assu}\label{asu1}
   Each entry of matrix $\bA$ is missing completely at random. Let $\delta_{u,i}$ denote i.i.d. Bernoulli random variable with parameter $\pi$. Let $\delta_{u,i}=1$ represents that the $a_{u,i}$ is observed and 0 otherwise.
\end{assu}
\begin{assu}\label{asu2}
  Matrix $\bZ$ is low rank such that $k \ll \min(m,n)$. The dimensions of $\bZ$ are at the same order such that $m/n=O(1)$.
\end{assu}
\begin{assu}\label{asu3}
  Squared singular values are bounded away from 0 such that $c_1^2 \geqslant c_2^2 \geqslant\cdots\geqslant c_k^2 \geqslant \eta_3 > 0$.
\end{assu}
\begin{assu}\label{asu4}
  Singular vectors are equally distributed. $p_{u,\ell}=O(1/\sqrt{n})$ for $\ell\in[k]$, $u\in[n]$ and $q_{\ell, i}=O(1/\sqrt{m})$ for $\ell\in[k]$, $i\in[m]$.
\end{assu}

Based on the singular value decomposition as shown in \ref{model_svd}, $\bx_{u}$ and $\by_{i}$ for a specific user $u$ and item $i$ can be defined as singular value decomposition scores as
\bea
    \bx_{u} = \bP_{(u)}\bTheta,\quad \by_{i} = \bTheta\bQ^{\T}_{i} \label{defxy},
\eea
where $\bx_u = (x_{u, 1}, \ldots, x_{u, k})$ and $x_{u, \ell} = p_{u, \ell} c_{\ell}$, $\by_i = (y_{1, i}, \ldots, y_{k, i})$ and $y_{t, i} = q_{\ell, i} c_{\ell}$. From the perspective of factor analysis, $\bx_u$ can be intuitively interpreted as latent factors for user $u$ and $\by_i$ as latent factors for item $i$. Followed by the above scores, we can write
\bea
    \bZ_{(u)} =& \bP_{(u)}\bTheta\bQ^{\T}
    = (p_{u,1}, \cdots, p_{u,k})\mbox{diag}(c_1, \cdots, c_k) [\bfq_1 \cdots \bfq_k]^{\T}
    = x_{u,1}\bfq_{1}^{\T}+\cdots+x_{u,k}\bfq_{k}^{\T}, \label{score1} \\
    \bZ_{i} =& \bP\bTheta\bQ_{i}^{\T}
    = [\bfp_1 \cdots \bfp_k] \mbox{diag}(c_1,\cdots,c_k)(q_{1,i}, \cdots, q_{r,i})^{\T}
    =\bfp_{1}y_{1,i}+\cdots+\bfp_{k}y_{k,i}. \label{score2}
\eea

Assume that each entry $z_{u,i}$ can be modeled by user-related characteristics $\bx_u \in \mathcal{X}$ and item-related features $\by_i \in \mathcal{Y}$ in a function based on the singular value decomposition scores above such that $g: \mathcal{X} \times \mathcal{Y} \rightarrow \mathbb{R}$. In addition, assume that $\mathcal{X}$ and $\mathcal{Y}$ are equipped with the Borel probability measure for $\bx_{u} \in \mathcal{X}$ representing the unobserved user-related features for a specific $u$ and $\by_{i}\in\mathcal{Y}$ the unobserved item-related features for a specific $i$. We can pool the latent factors in a function $g$ using the definitions as in equation \eqref{defxy} and it is shown as
\bea
    z_{u,i} = g(\bx_{u}, \by_i)=\sum_{\ell=1}^k\frac{1}{c_\ell}x_{u,\ell}y_{\ell,i}. \label{gfunc}
\eea

\subsection{Distance measure approximation}

We consider defining the distance measure between two latent feature vectors by
\bea
    d_{u,v} = \frac{1}{\sqrt{m}}\|\bx_u - \bx_v\|_2, & & d_{i,j} = \frac{1}{\sqrt{n}}\|\by_i - \by_j\|_2 \label{d},
\eea
where $\|\cdot\|_2$ denotes $L_2$ norm, to quantify the relationship between different users and items since it is insuperable to model the latent factors directly. The difficulty in obtaining the distance $d_{u,v}$ between two different users $u$ and $v$, and $d_{i,j}$ between two different items $i$ and $j$ lies in the hidden feature vectors $\bx_u$ and $\bx_v$, and $\by_i$ and $\by_j$. We then define $d_{u,v}^*$, $d_{i,j}^*$ and $\widetilde{d}_{u,v}$, $\widetilde{d}_{i,j}$ to turn to the entries within $\bA$ and $\bZ$ in the form of
\bea
    d^*_{u,v} = \frac{1}{\sqrt{m}}\|\bZ_{(u)} - \bZ_{(v)}\|_2, & &
    d^*_{i,j} = \frac{1}{\sqrt{n}}\|\bZ_{i} - \bZ_{j}\|_2, \label{dstar} \\
    \widetilde{d}_{u,v} =  \frac{1}{\sqrt{m}}\|\bA_{(u)} - \bA_{(v)}\|_2, & &
    \widetilde{d}_{i,j} =  \frac{1}{\sqrt{n}}\|\bA_{i} - \bA_{j}\|_2. \label{dtilde}
\eea

We can show that $d_{u,v}$ and $d^*_{u,v}$, $d_{i,j}$ and $d^*_{i,j}$ are equivalent by the expression of $\bZ$ using singular scores as in equations~\ref{score1} and \ref{score2}. For $\widetilde{d}_{u,v}$ and $d^*_{u,v}$, $\widetilde{d}_{i,j}$ and $d^*_{i,j}$, there exists non-negligible random error impact for the distance measure between the true underlying scores in $\bZ$ and the observed values in $\bA$, and we introduce the empirical variance estimator $\widehat{\sigma}^2$ derived from a first-step estimator $\widehat{z}^{(1)}_{v,j}$ as $\widehat{\sigma}^2=|\Omega|^{-1}\sum_{(v,j)\in\Omega}\big(a_{v,j}-\widehat{z}^{(1)}_{v,j}\big)^2$ to mitigate the discrepancy. To bridge the gap between the distance measure defined by latent factors and the distance measure defined by the observed values, we give the two propositions below.

\begin{prop}\label{prop4} \it Based on singular value decomposition of $\bZ$,
\bea
    \|\bZ_{(u)}-\bZ_{(v)}\|_2 = \|\bx_u-\bx_v\|_2, & & \|\bZ_{i}-\bZ_{j}\|_2 = \|\by_i-\by_j\|_2. \nn
\eea
\end{prop}

\begin{prop}\label{prop2} \it Under Assumption~\ref{asu1}, and assume that $|\widehat{\sigma}-\sigma| = O_p(m^{-1/2}n^{-1/2}\pi^{-1})$,
\bea
    \big | m^{-1}\|\bA_{(u)}-\bA_{(v)}\|_2^2 - 2\widehat{\sigma}^2 - m^{-1}\|\bZ_{(u)}-\bZ_{(v)}\|_2^2 \big | &=& O_p(m^{-1/2}n^{-1/2}\pi^{-1}), \nn \\
    \big | n^{-1}\|\bA_{i}-\bA_{j}\|_2^2 - 2\widehat{\sigma}^2 - n^{-1}\|\bZ_{i}-\bZ_{j}\|_2^2 \big | &=& O_p(m^{-1/2}n^{-1/2}\pi^{-1}). \nn
\eea
\end{prop} 

Proposition~\ref{prop4} states that the $L_2$ norm between two defined latent feature vectors is the same as the corresponding row- or column-wise $L_2$ norm in the true underlying score matrix $\bZ$. Proposition~\ref{prop2} states that the row- or column-wise squared $L_2$ norm in the observed matrix $\bA$ is the consistant estimator of those in the true underlying score matrix $\bZ$ after the introduction of the empirical variance estimator to adjust the noise. The convergence rate is the same as that for the empirical variance estimator. These two propositions make it valid for estimating the distance between latent feature vectors by the observed rating matrix. However, there exists one more obstacle to derive practical estimates from the observed rating matrix $\bA$, which is the high proportions of missingness. It could bring problems in the computation of $L_2$ norm between two rows of $\bA$, since some observed ratings may pair with an unobserved value. To address this, we choose only observed overlapped ratings within two rows and two columns to carry out the computation and get $\widehat{d}_{u,v}$ and $\widehat{d}_{i,j}$ as a realized distance approximation for $d_{u,v}$ and $d_{i,j}$. We first define $\mathcal{I}_u$ as the set of item indices that are observed to be scored by user $u$ and $\mathcal{U}_i$ as the set of user indices who are observed to rate item $i$ based on Assumption~\ref{asu1}:
\bea
\mathcal{I}_{u} = \{j: \delta_{u, j}=1 \ \mathrm{ for}\ j \in [m]\},& & 
\mathcal{U}_{i} = \{v: \delta_{v, i}=1 \ \mathrm{ for}\ i \in [n]\}, \nn
\eea
then the intersection between two ``item sets" or ``user sets", i.e., the common items rated by user $u$ and $v$ and the common users rate both item $i$ and $j$, can be denoted as $\mathcal{I}_{uv}=\mathcal{I}_u\cap\mathcal{I}_v$ and $\mathcal{U}_{ij}=\mathcal{U}_i\cap\mathcal{U}_j$, correspondingly. We can therefore write $\widehat{d}_{u,v}$ and $\widehat{d}_{i,j}$ as
\bea
    \widehat{d}_{u,v}^2 =
    \frac{1}{|\mathcal{I}_{uv}|}
    \sum_{i \in \mathcal{I}_{uv}} (a_{u, i} - a_{v, i})^2, & & 
    \widehat{d}_{i,j}^2 =
    \frac{1}{|\mathcal{U}_{ij}|}
    \sum_{u \in \mathcal{U}_{ij}} (a_{u, i} - a_{u, j})^2 \label{dhat},
\eea
where $|\cdot|$ denotes the size of a set, and with convention that the score is null when the corresponding intersection set is empty. We give proposition~\ref{prop3} to show the relationship between $\widehat{d}_{u,v}$ and $\widetilde{d}_{u,v}$, $\widehat{d}_{i,j}$ and $\widetilde{d}_{i,j}$ and establish that the practical estimator $\widehat{d}_{\cdot, \cdot}$ which relies on the actual observed values considering missingness is consistent with the theoretical fully observed rating matrix.

\begin{prop}\label{prop3} \it Under Assumption~\ref{asu1},  we can have
\bea
    \big|\widehat{d}^2_{u,v} - m^{-1}\|\bA_{(u)}-\bA_{(v)}\|^2_2 \big| &=& O_p(m^{-1/2}\pi^{-1}), \nn \\
    \big|\widehat{d}^2_{i,j} - n^{-1}\|\bA_{i}-\bA_{j}\|^2_2 \big| &=& O_p(n^{-1/2}\pi^{-1}), \nn
\eea
for the model defined in equation~\ref{model}.
\end{prop}

In summary, we provide Theorem~\ref{theorem3} to connect the distance measure that is computable from $\bA$ and the theoretical distance measure derived from latent feature vectors

\begin{theo}\label{theorem3}
Under Assumption~\ref{asu1}, and assume that $|\widehat{\sigma}-\sigma| = O_p(m^{-1/2}n^{-1/2}\pi^{-1})$, we can have
\bea
\big|\widehat{d}_{u,v}^2 - 2\widehat{\sigma}^2 - d_{u,v}^2 \big| &=&  O_p(m^{-1/2}n^{-1/2}\pi^{-1}), \\
\big|\widehat{d}_{i,j}^2 - 2\widehat{\sigma}^2 - d_{i,j}^2 \big| &=&  O_p(m^{-1/2}n^{-1/2}\pi^{-1}).
\eea
\end{theo}

Despite $\bx_{u}$ and $\by_i$ being unobservable, we are able to estimate the distance in their latent feature space by leveraging the actual observed ratings from the corresponding rows and columns of the available observed matrix. This also explains the rationality behind the classical collaborative filtering, where the imputations are built on the weighted average of some other available ratings through the row-wise or column-wise correlation.

\subsection{Radial neighbors and smoothing estimation}

Using the defined distance, we propose employing kernel regression to approximate the unobserved function $g$ in \eqref{gfunc}. The first step involves defining a neighboring set for information pooling. A straightforward approach is to select sets based on row-wise or column-wise correlations, which necessitates at least two overlapping observed values for the same user or item. However, this approach has a major limitation that feasible sets are sparse due to insufficient common ratings. Additionally, it can be challenging to address reference ratings that exhibit negative or insignificant correlations.

To exploit and incorporate more information within the original matrix, we propose radial neighbors to represent all the cells worth information pooling directly or indirectly. The intuition is that, in addition to $a_{vi}$ for the item of interest $i$, all other scores $a_{vj}$ for $j\neq i$ would also provide ``indirect'' information through the rating history of user $v$ and the item relationship of item $j$. The rationale for defining radial neighbors is to collect all ratings with which an approximate $L_2$ norm can be computed in the latent user or item feature space with the target user-item pair. We define the radial neighbor set as follows

\bea
\mathcal{N}_{ui}^{\mathrm{Rdl}} &=& \{(v,\cdot): (v,\cdot) \in \Omega, ~|\mathcal{I}_{uv}| \geqslant\beta\} \cup \{(\cdot,j): (\cdot, j) \in \Omega, ~|\mathcal{U}_{ij}| \geqslant\beta\}. \label{radial_neighbor1}
\eea

We use a toy example to explain the construction of radial neighbor set. Assume an observed rating matrix with $m=n=5$ as shown in formula~\ref{toyexample1} where missing values are left blank. We can write the rating set $\Omega=\{a_{1,1}, a_{1,3}, a_{2,2}, a_{3,1}, a_{3,4}, a_{4,3}, a_{4,5}, a_{5,2}, a_{5,4}\}$. If our goal is to estimate the missing $a_{1,4}$, we can first reorganize the matrix layout as in the format shown in formula~\ref{toyexample2} by concatenating the observed ratings in row and in column to match the subscript. The vertical dashed line separates the row- and column-wise observed ratings, and the two horizontal dashed lines surround the missing rating $a_{1,4}$ we want to estimate. Assume $\beta=1$ and then we can identify $v=\{1,3,4\}$ and $j=\{1,2,4\}$ by checking $\mathcal{I}_{1v}\geqslant 1$ and $\mathcal{U}_{4j}\geqslant 1$. We can construct the radial neighbor set for $a_{1,4}$ as $\mathcal{N}_{14}^{\mathrm{rdl}}=\{a_{1,1},a_{1,3}, a_{3,1}, a_{3,4}, a_{4,3},a_{4,5}, a_{2,2}, a_{5,2}, a_{5,4}\}$, which covers all observed ratings.
\bea
\label{toyexample1}
\left(\begin{array}{cccccc} & \text { item1 } & \text { item2 } & \text { item3 } & \text { item4 } & \text { item5 } \\ \text { user1 } & a_{1,1} & & a_{1,3} & ? & \\ \text { user2 } & & a_{2,2} & & & \\ \text { user3 } & a_{3,1} & & & a_{3,4} & \\ \text { user4 } & & & a_{4,3} & & a_{4,5} \\ \text { user5 } &  & a_{5,2} & & a_{5,4} & \end{array}\right)
\eea

\bea
\label{toyexample2}
\left(\begin{array}{l}
a_{1,1} \\
a_{1,3} \\
\hdashline
a_{1,4} \\
\hdashline
a_{2,2} \\
a_{3,1} \\
a_{3,4} \\
a_{4,3} \\
a_{4,5} \\
a_{5,2} \\
a_{5,4}
\end{array}\right) \sim\left(\begin{array}{lllll:lllll}
a_{1,1} & & a_{1,3} & &  & a_{1,1} & & a_{3,1} & & \\
a_{1,1} & & a_{1,3} & & & a_{1,3} & & & a_{4,3} & \\
\hdashline
a_{1,1} & & a_{1,3} & & & & & a_{3,4} & & a_{5,4} \\
\hdashline
& a_{2,2} & & & & & a_{2,2} & & & a_{5,2} \\
a_{3,1} & & & a_{3,4} & & a_{1,1} & & a_{3,1} & & \\
a_{3,1} & & & a_{3,4} & & & & a_{3,4} & & a_{5,4}\\
& & a_{4,3} & & a_{4,5} & a_{1,3} & & & a_{4,3} & \\
& & a_{4,3} & & a_{4,5} & & & & a_{4,5} & \\
& a_{5,2} & & a_{5,4} & & & a_{2,2} & & & a_{5,2} \\
& a_{5,2} & & a_{5,4} & & & & a_{3,4} & & a_{5,4} \\
\end{array}\right)
\eea

With the radial neighbors at hand, we proposed radial neighborhood estimator (RNE) by adopting the framework of Nadaraya-Waston estimator which performs locally kernel regression to smooth the observations within the radial neighbor set, where the kernel function can hold a joint distribution of user features related distance and item features related distance. It is noted that using $L_2$ norm within kernel can avoid the problem of unknown dimensions of the latent feature spaces.
\bea
\widehat{z}_{u, i}
=
\widehat{g}(\bx_{u}, \by_i)
&=&
\sum_{(v, j) \in \mathcal{N}_{ui}^{\mathrm {Rdl }}} 
\frac{ K\big(\frac{\|\bx_{v} - \bx_{u}\|_2}{h_1}, 
                \frac{\|\by_{i} - \by_{j}\|_2}{h_2}\big)}
      {\sum_{(v, j) \in \mathcal{N}_{ui}^{\mathrm {Rdl }}} K\big(\frac{\|\bx_{v} - \bx_{u}\|_2}{h_1}, \frac{\|\by_{i} - \by_{j}\|_2}{h_2}\big) } z_{v,j}.
\eea

The distance between the bivaraite unknown features above can be approximated by row- and column-wise euclidean distances based on Theorem~\ref{theorem3}, and $2\widehat{\sigma}^2$ is to offset the bias that arises from the squared terms involving the observation noise. The observed ratings $a_{v,j}$ are incorporated into the final practical RNE $\widetilde{z}_{u,i}$, and we will discuss the consistency of this approach in greater detail in Section~\ref{section3}.
\bea
\widetilde{z}_{u,i}
&=&
    \sum_{(v, j) \in \mathcal{N}_{ui}^{\mathrm {Rdl }}}
    \frac{K\big(\frac{\sqrt{\widehat{d}_{u,v}^2-2\widehat{\sigma}^2}}{h_1}, 
                \frac{\sqrt{\widehat{d}_{i,j}^2-2\widehat{\sigma}^2}}{h_2}\big)}
        {\sum_{(v, j) \in \mathcal{N}_{ui}^{\mathrm {Rdl }}} K\big(\frac{\sqrt{\widehat{d}_{u,v}^2-2\widehat{\sigma}^2}}{h_1}, 
                \frac{\sqrt{\widehat{d}_{i,j}^2-2\widehat{\sigma}^2}}{h_2}\big) } a_{v,j} \label{z_tilde}.
\eea

\subsection{Related methods}
There are several methods that are relatively closely related to our proposed approach, which we intend to compare here. SoftImpute \citep{mazumder2010spectral} is a iterative soft-threholded SVD for matrix completion using nuclear norm regularization and has contributed much to the matrix factorization based approaches. The criterion considered is 
\bea
    \min _{Z} && \frac{1}{2}\left\|P_{\Omega}(\bA)-P_{\Omega}(\bZ)\right\|_{F}^{2}+\lambda\|\bZ\|_{*},
\eea
where the projection $P_{\Omega}(\bX)$ is the matrix with observed elements of $\bX$ preserved, and the unobserved entries replaced with 0, $\|\bZ\|_*$ denotes the nuclear norm of $\bZ$. The solution is proved to be $\widehat{\bM}=S_{\lambda}(\bZ)$, where the operator $S_{\lambda}(\bZ)$ is a factorization of $\bZ=\bP\bTheta\bQ^T$ and then a reconstruction $S_{\lambda}(\bZ)=\bP\bTheta^*\bQ^T$ for the ``filled-in" version of $\bZ$ using the soft-thresholded singular values $d_{i}^{*}=\mathbb{I}\left(c_{i}-\lambda>0\right)$.

The neighbor based approaches are featured by collaborative filtering which was proposed in \citep{goldberg1992using} and have become practical due to its simplicity and ability to scale \citep{desrosiers2011comprehensive}. The key to prediction by collaborative filtering is to pool the ratings from either the users or items that share preference similarities with the target one. We present the model using the framework for user-based collaborative filtering and it can be easily generalized to item-based version
\bea
    \widehat{a}_{u,i} = \frac{1}{|\mathcal{N}_{u}|}\sum_{v\in\mathcal{N}_{u}}a_{v,i},~~ \mathrm{ or } ~~
    \widehat{a}_{u,i} = \frac{\sum_{v\in\mathcal{N}_{u}}w_{u,v}a_{v,i}}{\sum_{v\in \mathcal{N}_{u} |w_{u,v}|}}
\eea
with some properly-defined similarity measures $w_{u,v}$. The commonly-picked ones include cosine similarity \citep{balabanovic1997fab, billsus2000user}, Pearson correlation similarity, Spearman rank correlation similarity, etc. $\mathcal{N}_u\in\Omega$ denotes the indices that can be viewed as "neighbors" for user $u$. For a specific rating for item $i$ by user $u$ imputed by user-based collaborative filtering, the available references ratings $a_{v,i}$ for $v\in[m], v\neq u$ are first required. At least two more extra common ratings should be present for $u$ and any $v$ so that can correlation be computed by possible means. Thus we can write neighbouring set for user-based collaborative filtering as $\mathcal{N}_{u}=\{(v,i):(v,i)\in\Omega \mathrm{~for~} i\in\mathcal{I}_v, \mathcal{I}_{uv}\backslash\{i\}\neq\emptyset\}$.

Several recent neighborhood-based approaches caught our attention. Blind regression established the framework on a nonparameteric regression over latent variable models inspired by Taylor's expansion \citep{song2016blind}. The estimator is defined as
\bea
    \widehat{a}_{u,i} &=& \frac{\sum_{(v,j)\in B^{\beta}(u,i)}w_{u,i}(v,j)(a_{u,j}+a_{v,i}-a_{v,j})}{\sum_{(v,j)\in B^{\beta}(u,i)}w_{u,i}(v,j)}.
\eea
The set $B^{\beta}(u,i)$ denote the cells $(v,j)$ within the matrix such that the entries $a_{v,j}, a_{u,j}$ and $a_{v,i}$ are all observed, and the commonly observed ratings between $(u,v)$ and between $(i,j)$ are at least $\beta$. $\beta$ as a parameter controls the degree of observed ratings overlap between rows or columns and therefore affects the accuracy of the calculation for empirical variance $s_{uv}^2$ and $s_{ij}^2$, which is designed as the similarity measure in this work. Compared to other neighbor-based methods including collaborative filtering or blind regression \cite{song2016blind}, it is intuitive that the sets of neighbors defined in these methods are subsets of the set consisting of radial neighbors, since distance measures such as correlation coefficients or empirical variance require at least two simultaneously observed common ratings, while the $L_2$ norm requires only one.

\cite{yu2022nonparametric} proposed a nonparametric smoothing method that utilizes one-sided covariates associated with either the rows or the columns of the matrix to compute the euclidean distance between entities based on their covariate vectors. This method is similar to our proposed approach in the paper, but we relax the constraints to address situations where there are no row or column covariates.



\section{Theoretical results}\label{section3}

In this section, we provide the main theorems to show that the proposed estimator is consistent and rigorous. More technical details can be found in Section~\ref{appendix}.

We first establishes that the function $g$, which maps user-item pairs to latent space distances, is $L$-Lipschitz with respect to the $\infty$-product metric in Lemma~\ref{lemma1}. Specifically, it ensures that the variation in $g$ is bounded by the maximum of scaled user and item distances in the latent space, guaranteeing that small changes in the latent space lead to proportionally small variations in score distances. The parameter $L$ is related to the number of non-zero singular values. In Lemma~\ref{lemma2}, we show the concentration of the size of radial neighbor set $ |\mathcal{N}_{ui}^{\mathrm{Rdl}}|$. For a given pair $(u,i)\in [n] \times [m]$, there exists a feasible radial neighbor set and the number of radial neighbors concentrates to $mn\pi^2$ with high probability. In Lemma~\ref{lemma3}, we write the estimated density function for the latent feature vectors $(\bx_{u},\by_i)$ and its corresponding radial neighbor set $\mathcal{N}_{ui}^{\mathrm{Rdl}}$ in kernel regression. We can show that the estimated density function converges to its expected value with probability of one when the variance vanishes asymptotically given the radial neighbor set $\mathcal{N}_{ui}^{\mathrm{Rdl}}$ growing to infinity and appropriate bandwidth $h_1$ and $h_2$. The useful Lemma~\ref{lemma3} assists the subsequent proof of the consistency of the proposed estimator.

\begin{lem}\label{lemma1} 
Under Assumptions~\ref{asu2}, \ref{asu3}, \ref{asu4}, function $g$ is $L-$Lipschitz with respect to $\infty$-product metric when $L=4\sqrt{k}$ as
\bea
|z_{u,i}-z_{v,j}|=|g(\bx_{u}, \by_i)-g(\bx_{v}, \by_j)| \leqslant L \cdot  \{m^{-1/2}\|\bx_u-\bx_v\|_2 \wedge n^{-1/2}\|\by_i-\by_j\|_2\}. \nn
\eea
Here $\{a \wedge b\} = \max\{ a,b\} $.
\end{lem}

\begin{lem}\label{lemma2}
Given $\alpha\in(0,1)$, for any pair $(u,i)\in [n]\times[m]$,
\bea
    \mathbb{P}\big( |\mathcal{N}_{ui}^{\mathrm{Rdl}}| \notin (1 \pm \alpha)mn\pi^2\big) 
    &\leqslant&
    2m\exp \bigg(-\frac{\alpha^2}{3}n\pi^2\bigg) \wedge 2n\exp \bigg(-\frac{\alpha^2}{3}m\pi^2\bigg). \nn
\eea
\end{lem}

\begin{lem}\label{lemma3}
Define $\widetilde{f}(\bx_{u},\by_i)$ as a pseudo form of estimated density function for $(\bx_u, \by_i)$ as
\bea
\widetilde{f}(\bx_{u},\by_i)&=& \frac{1}{h_1h_2|\mathcal{N}_{ui}^{\mathrm{rdl}}|}
\sum_{(v,j)\in\mathcal{N}_{ui}^{\mathrm{rdl}}}
K\bigg(\frac{\|\bx_{v}-\bx_{u}\|_2}{h_1},
        \frac{\|\by_j-\by_i\|_2}{h_2}\bigg) \nn
\eea
and under Assumptions~\ref{asu1}, $|\mathcal{N}_{ui}^{\mathrm{Rdl}}|=O_p(m^{-1}n^{-1}\pi^{-2})$, $h_1=h_2=O_p(m^{-1/2}n^{-1/2}\pi^{-1})$, we have
\bea
\frac{\widetilde{f}(\bx_u, \by_i)}{\mathbb{E}\{\widetilde{f}(\bx_u, \by_i)\}}\stackrel{p}{\rightarrow} 1 \nn
\eea
\end{lem}

We then establish the consistency of the RNE $\widehat{z}_{u,i}$ in Theorem~\ref{theorem1}, showing that it converges to the true latent value $z_{u,i}$ for a given user-item pair $(u,i)$. The consistency hold under feasible conditions that the radial neighbor set is at the order of the inverse of the number of observed ratings and that the corresponding kernel regression bandwidths $h_1$ and $h_2$ are controlled by the square root of the number of observed ratings.  In Theorem~\ref{theorem2}, we further prove the consistency of practical RNE $\widetilde{z}_{u,i}$, which is the computable form of the proposed estimator derived from the actual observed rating matrix $\bA$, to the proposed estimator $\widehat{z}_{u,i}$. This two-step proof framework effectively bridges the theoretical gap between the unobserved true rating $z_{u,i}$ in $\bZ$ and the practical RNE rating $\widetilde{z}_{u,i}$ in $\bA$, ensuring the robustness of our approach. Note that general conditions for the kernel function are required including $K$ is a symmetric function, $\iint K(a,b) = 1$, $\iint K^2(a,b)\leqslant \infty$, $\iiint K(a,b)K(a',b) \leqslant \infty$, $\iiiint K(a,b)K(a',b') \leqslant\infty$.

\begin{theo}\label{theorem1}
Under Assumptions~\ref{asu1}, \ref{asu2}, \ref{asu3}, \ref{asu4}, and assume $|\widehat{\sigma}-\sigma| = O_p(m^{-1/2}n^{-1/2}\pi^{-1})$, function $g$ is $L$-Lipschitz continuous, $|\mathcal{N}_{ui}^{\mathrm{Rdl}}|=O_p(m^{-1}n^{-1}\pi^{-2})$, $h_1=h_2=O_p(m^{-1/2}n^{-1/2}\pi^{-1})$
, it can be obtained that $\widehat{z}_{u,i} \to z_{u,i}$.
\end{theo}

\begin{theo}\label{theorem2}
Under Assumption~\ref{asu1}, and assume $|\widehat{\sigma}-\sigma| = O_p(m^{-1/2}n^{-1/2}\pi^{-1})$, $h_1=h_2=O_p(m^{-1/2}n^{-1/2}\pi^{-1})$, it can be obtained that $\widetilde{z}_{u,i} \to \widehat{z}_{u,i}$.
\end{theo}

\section{Numerical results}\label{section4}

In this section, we compared the performances of several relative methods, including collaborative filtering, soft impute, blind regression, together with RNE on both generated datasets and real-world cases. We used independent standard Gaussian kernel weights in RNE. $\beta=1$ was chosen here to pool as many radial neighbors as possible. The bandwidths and the weight combination were optimized by 5-fold cross validation on combinatorial grid search to reach the minimum RMSE. For blind regression, normal kernel weight $w_{u i}(v, j)=\exp \big(-\lambda \min \big\{s_{u v}^{2}, s_{i j}^{2}\big\}\big)$, $\lambda=\{0.001, 0.01, 0.1, 1, 2, 3\}$, and $\beta=2$ were used. For softImpute, regularization $\lambda$ was tried from 0.1 to 4.1 and the step size was 0.5. Both of the parameters $\lambda$ in blind regression and soft impute were tuned by 5-fold cross validation. For collaborative filtering, the distance metric was set as euclidean distance and only neighbors with positive correlations were used.

\subsection{Simulation study}

In all our simulations, we assumed the underlying model $Z_{n\times m}=U_{m\times k}V_{k\times n}'+\epsilon=Z_0+\epsilon$ where $U$ and $V$ were random matrices with standard normal Gaussian entries. We used $Z_0$ to denote the truth part, and $\epsilon$ is i.i.d. Gaussian with mean 0 and variance $\sigma^2_{\epsilon}$. The $\sigma^2_{\epsilon}$ was determined by the signal to noise ratio (SNR), which is defined as $\mathrm{SNR}=\sqrt{\mathrm{Signal}(Z_0)/\sigma^2_{\epsilon}}$ where $\mathrm{signal}(Z_0)=\sum_{i=1}^n\sum_{j=1}^m(Z_{0ij}-\bar{Z}_0)^2/(nm-1)$ and $\bar{Z}_0=\sum_{i=1}^n\sum_{j=1}^mZ_{0ij}/(nm)$. We chose $\mathrm{SNR}=1$ in our simulations. 

The missing probability denoted as $p$ was applied uniformly to rows and columns of the matrix. Besides, we introduced a ``cold-start" rate $\phi$ to mimic the real-world case to examine the prediction accuracy of the experimented models: after filtering out those entries that were labeled as ``missing", we randomly selected $\phi$ proportion of row indices (i.e., users) and column indices (i.e., items) and retain all of their observed entries in the testing set. The remaining observed entries were then randomly split to ultimately form the training set $(\Omega_{\mathrm{train}})$ and testing set $(\Omega_{\mathrm{test}})$ with 75\% and 25\% of the total observed entries. We used three empirical measures to evaluate the accuracy and coverage of the matrix completion.
\bea
    \mathrm{RMSE} &=& \frac{\|\hat{Z}_{\Omega_{\mathrm{test}}}-Z_{\Omega_{\mathrm{test}}}\|_F}{\sqrt{|\Omega_{\mathrm{test}}|}} \nn \\
    \mathrm{standardized~test~error} &=& \frac{\|Z_{0,\Omega_{\mathrm{test}}}-\hat{Z}_{\Omega_{\mathrm{test}}}\|_F^2}{\|Z_{0,\Omega_{\mathrm{test}}}\|_F^2} \nn \\
    \mathrm{NA~value~proportion} &=& \frac{|\{x\in \Omega_{\mathrm{test}} | x = N.A.\}|}{|\Omega_{\mathrm{test}}|} \nn
\eea

Since collaborative filtering, soft impute, and blind regression cannot handle cold-start cases, we further split the test set in order to be able to do a fair comparison: $\Omega_{\mathrm{test}}=\Omega_{\mathrm{test, cold-start}}+\Omega_{\mathrm{test, non-cold-start}}$, where $\Omega_{\mathrm{test, cold-start}}$ contained the entries with either row indices (user) or column indices (item) not observed in the training set and $\Omega_{\mathrm{test, non-cold-start}}$ contained entries whose row and/or column indices existed in the training set. All the aforementioned methods were applied on $\Omega_{\mathrm{test, non-cold-start}}$ for evaluation and RNE was also applied on $\Omega_{\mathrm{test, cold-start}}$. Table~\ref{tab:sim_rmse_100} and \ref{tab:sim_rmse_300} showed RMSE comparisons on $\Omega_{\mathrm{test, non-cold-start}}$ for $m=n=100$ and $m=n=300$. Table~\ref{tab:sim_na_100} and \ref{tab:sim_na_300} showed NA value proportion comparisons on $\Omega_{\mathrm{test}}$ for $m=n=100$ and $m=n=300$. Table~\ref{tab:sim_error_100} and 6 showed standardized test error comparisons on $\Omega_{\mathrm{test, non-cold-start}}$ for $m=n=100$ and $m=n=300$. Table~\ref{tab:sim_coldstart} showed the performance of RNE on $\Omega_{\mathrm{test, cold-start}}$. Each simulation was repeated for 100 times and we outputted mean and standard error values. Those values representing the best performance are in bold in the tables.

From Table~\ref{tab:sim_rmse_100} and \ref{tab:sim_rmse_300}, it can be concluded that RNE outperformed the other methods especially when the missing probability and/or ``cold-start" rate is high in both dimension settings of 100 and 300. The rank also influenced the performance: collaborative filtering and soft impute could produce better results when the rank is extremely low, i.e., $k=3$ when $m=n=300$ or $m=n=100$ and $k=6$ when $m=n=300$. However, the performances of these tow methods would be compromised with the increasing missing probability and ``cold-start" rate. Similar patterns could also be observed in Table~\ref{tab:sim_error_100} and \ref{tab:sim_error_300} about standardized test error. It could be noted that low dimension ($m=n=100$) possibly provided less neighboring information such that the results were more biased compared to those given high dimension ($m=n=300$). 

Table~\ref{tab:sim_na_100} and \ref{tab:sim_na_300} showed clearly that RNE maintained a rather low proportion of NA values regardless of dimension, rank, missing probability, and ``cold-start" rate. Intuitively, soft impute can be used as a benchmark to detect the presence of row index or column index that are neither not observed in the training set as it can only complete the matrix with original size. This dual ``cold-start" accounted for the values reported in Table~\ref{tab:sim_na_100} and \ref{tab:sim_na_300} for RNE and these cases remain a challenge to RNE as there is no neighboring information to borrow.

Table~\ref{tab:sim_coldstart} showed the results of RNE on $\Omega_{\mathrm{test, cold-start}}$ so only those given $\phi=0.05$ and $0.1$ were given. All the three empirical metrics increased a bit compared to the results produced by $\Omega_{\mathrm{test, non-cold-start}}$ and this indicated the existence of bias introduced by the lack of neighboring information from either row or column space.

\begin{table}[!htbp]
\centering
\begin{tabular}{llllll}
\hline
$m=n=100$   & CFuser & CFitem & softImpute & blind reg  & RNE\\ \hline
$\pi=0.6,\ k=3$  & \multicolumn{5}{l}{}  \\ 
$\phi=0$ & 3.42(.043) & 3.41(.043) & 3.47(.047) & 3.46(.042)  & \bftab 3.40(.040)    \\ 
$\phi=0.05$ & \bftab3.39(.041) &  3.39(.042) & 3.43(.044) & 3.44(.041)  & 3.40(.039)  \\ 
$\phi=0.1$ & 3.38(.041) & \bftab3.34(.041) & 3.37(.043) & 3.40(.039)   & 3.38(.038)  \\ \hline
\end{tabular}
\begin{tabular}{llllll}
\hline
$\pi=0.6,\ k=6$  & \multicolumn{5}{l}{}  \\ 
$\phi=0$ & 6.65(.052) & 6.67(.052) & 7.31(.060) & 6.61(.052)  & \bftab6.41(.050)    \\ 
$\phi=0.05$ & 6.61(.057) & 6.61(.056) & 7.15(.066) & 6.54(.054)  & \bftab6.39(.054) \\ 
$\phi=0.1$ & 6.54(.059) & 6.52(.060) & 7.10(.065) & 6.48(.058)  & \bftab6.32(.058)    \\ \hline
\end{tabular}

\begin{tabular}{llllll}
\hline
$\pi=0.6, k=10$  & \multicolumn{5}{l}{}   \\ 
$\phi=0$ & 10.8(.072) & 10.8(.072) & 11.9(.092) & 10.7(.073)  & \bftab10.3(.072)   \\ 
$\phi=0.05$ & 10.7(.074) & 10.7(.074) & 11.8(.081) & 10.6(.072)  & \bftab10.3(.070)    \\ 
$\phi=0.1$ & 10.7(.071) & 10.7(.076) & 11.9(.091) & 10.6(.075)   & \bftab10.3(.071)    \\ \hline
\end{tabular}

\begin{tabular}{llllll}
\hline
$\pi=0.9,\ k=3$  & \multicolumn{5}{l}{} \\ 
$\phi=0$ & 4.42(.058) & 4.38(.061) & 4.20(.054) & 5.97(.086)  & \bftab3.48(.042)  \\ 
$\phi=0.05$ & 4.47(.063) & 4.56(.069) & 4.21(.059) & 5.73(.085) & \bftab3.45(.044)    \\ 
$\phi=0.1$ & 4.43(.079) & 4.46(.085) & 4.26(.059) & 5.61(.120)   & \bftab3.50(.053)    \\ \hline
\end{tabular}

\begin{tabular}{llllll}
\hline
$\pi=0.9,\ k=6$  & \multicolumn{5}{l}{} \\ 
$\phi=0$ & 8.67(.102) & 8.63(.099) & 8.19(.075) & 11.4(.154) & \bftab6.54(.059)  \\ 
$\phi=0.05$ & 8.54(.125) & 8.45(.114) & 8.24(.096) & 10.8(.186) & \bftab6.56(.066)    \\ 
$\phi=0.1$ & 8.22(.145) & 8.22(.162) & 8.40(.086) & 10.6(.173)   & \bftab6.45(.061)    \\ \hline
\end{tabular}

\begin{tabular}{llllll}
\hline
$\pi=0.9, k=10$  & \multicolumn{5}{l}{} \\ 
$\phi=0$ & 13.6(.140) & 14.0(.090) & 13.7(.129) & 18.1(.151)  & \bftab10.6(.079)  \\ 
$\phi=0.05$ & 13.9(.151) & 13.5(.149) & 13.9(.120) & 17.8(.185)  & \bftab10.6(.088)    \\ 
$\phi=0.1$ & 13.2(.167) & 13.1(.204) & 13.7(.164) & 17.7(.258)   & \bftab10.5(.104)    \\ \hline
\end{tabular}
\caption{RMSE for models on the test set without ``cold-start" entries. The dimension for the simulated matrix is $m=n=100$.}
\label{tab:sim_rmse_100}
\end{table}
\begin{table}[!htbp]
\centering
\begin{tabular}{llllll}
\hline
$m=n=300$   & CFuser & CFitem & softImpute & blind reg & RNE\\
\hline
$\pi=0.6,\ k=3$  & \multicolumn{5}{l}{} \\ 
$\phi=0$ & 3.22(.018) & 3.22(.018) & \bftab3.20(.017) & 3.34(.017)  & 3.42(.017)  \\ 
$\phi=0.05$ & 3.20(.019) & 3.20(.019) & \bftab3.19(.017) & 3.33(.018)  & 3.41(.019)    \\ 
$\phi=0.1$ & 3.21(.018) & 3.21(.018) & \bftab3.20(.017) & 3.34(.016)  & 3.43(.017)   \\ \hline
\end{tabular}

\begin{tabular}{llllll}
\hline
$\pi=0.6,\ k=6$  & \multicolumn{5}{l}{} \\ 
$\phi=0$ & 6.43(.032) & \bftab6.43(.031) & 6.52(.032) & 6.48(.031) & 6.44(.031)  \\ 
$\phi=0.05$ & \bftab6.47(.031) & 6.47(.032) & 6.56(.032) & 6.53(.031)  & 6.49(.031)    \\ 
$\phi=0.1$ & 6.31(.029) & \bftab6.30(.030) & 6.39(.032) & 6.38(.030) & 6.34(.029)    \\ \hline
\end{tabular}

\begin{tabular}{llllll}
\hline
$\pi=0.6, k=10$  & \multicolumn{5}{l}{} \\ 
$\phi=0$ & 10.6(.042) & 10.6(.041) & 10.9(.043) & 10.5(.041) & \bftab10.4(.041)   \\ 
$\phi=0.05$ & 10.6(.044) & 10.6(.044) & 10.9(.047) & 10.5(.043)  & \bftab10.4(.043)    \\ 
$\phi=0.1$ & 10.5(.038) & 10.5(.037) & 10.8(.036) & 10.5(.038)  & \bftab10.4(.037)    \\ \hline
\end{tabular}

\begin{tabular}{llllll}
\hline
$\pi=0.9,\ k=3$  & \multicolumn{5}{l}{} \\ 
$\phi=0$ & 3.79(.019) & 3.80(.019) & 3.68(.021) & 3.89(.020) & \bftab3.45(.017)  \\ 
$\phi=0.05$ & 3.71(.022) & 3.73(.021) & 3.66(.023) & 3.77(.023) & \bftab3.44(.020)    \\ 
$\phi=0.1$ & 3.67(.019) & 3.68(.022) & 3.64(.022) & 3.71(.020)   & \bftab3.46(.019)    \\ \hline
\end{tabular}

\begin{tabular}{llllll}
\hline
$\pi=0.9,\ k=6$  & \multicolumn{5}{l}{} \\ 
$\phi=0$ & 7.22(.031) & 7.23(.031) & 7.50(.034) & 7.35(.032)  & \bftab6.49(.027)  \\ 
$\phi=0.05$ & 7.08(.035) & 7.06(.034) & 7.49(.039) & 7.13(.034)  & \bftab6.47(.031)    \\ 
$\phi=0.1$ & 6.95(.034) & 6.97(.039) & 7.51(.042) & 7.00(.037)   & \bftab6.50(.034)    \\ \hline
\end{tabular}

\begin{tabular}{llllll}
\hline
$\pi=0.9, k=10$  & \multicolumn{5}{l}{} \\ 
$\phi=0$ & 11.8(.046) & 11.8(.050) & 12.4(.046) & 12.2(.053)  & \bftab10.6(.043)  \\ 
$\phi=0.05$ & 11.6(.046) & 11.6(.047) & 12.4(.049) & 11.8(.049)  & \bftab10.6(.043)    \\ 
$\phi=0.1$ & 11.4(.057) & 11.3(.058) & 12.4(.068) & 11.5(.058)   & \bftab10.6(.054)    \\ \hline
\end{tabular}
\caption{RMSE for models on the test set without ``cold-start" entries. The dimension for the simulated matrix is $m=n=300$.}
\label{tab:sim_rmse_300}
\end{table}
\begin{table}[!htbp]
\centering
\begin{tabular}{llllll}
\hline
$m=n=100$   & CFuser & CFitem & softImpute & blind reg & RNE \\ \hline
$\pi=0.6,\ k=3$  & \multicolumn{5}{l}{}    \\ 
$\phi=0$ & \bftab.000(.000) & \bftab.000(.000) & \bftab.000(.000) & \bftab.000(.000) & \bftab.000(.000)   \\ 
$\phi=0.05$ & .392(.001) & .392(.001) & .392(.001) & .392(.001) & \bftab.010(.000)  \\ 
$\phi=0.1$ & .764(.002) & .764(.002) & .764(.002) & .764(.002) & \bftab.041(.000)   \\ \hline
\end{tabular}

\begin{tabular}{llllll}
\hline
$\pi=0.6,\ k=6$  & \multicolumn{5}{l}{}  \\ 
$\phi=0$ & \bftab.000(.000) & \bftab.000(.000) & \bftab.000(.000) & \bftab.000(.000) & \bftab.000(.000)   \\ 
$\phi=0.05$ & .388(.001) & .388(.001) & .388(.001) & .388(.001) & \bftab.010(.000)  \\ 
$\phi=0.1$ & .759(.002) & .759(.002) & .759(.002) & .759(.002)  & \bftab.039(.000)   \\ \hline
\end{tabular}

\begin{tabular}{llllll}
\hline
$\pi=0.6, k=10$  & \multicolumn{5}{l}{}  \\ 
$\phi=0$ & \bftab.000(.000) & \bftab.000(.000) & \bftab.000(.000) & \bftab.000(.000) & \bftab.000(.000)   \\ 
$\phi=0.05$ & .392(.001) & .392(.001) & .392(.001) & .392(.001) & \bftab.009(.000)  \\ 
$\phi=0.1$ & .764(.002) & .764(.002) & .764(.002) & .764(.002)  & \bftab.040(.000)   \\ \hline
\end{tabular}

\begin{tabular}{llllll}
\hline
$\pi=0.9,\ k=3$  & \multicolumn{5}{l}{}  \\ 
$\phi=0$ & .709(.003) & .709(.003) & \bftab.000(.000) & .664(.004) & \bftab.000(.000)   \\
$\phi=0.05$ & .766(.003) & .764(.002) & .389(.004) & .735(.003) & \bftab.010(.001)  \\ 
$\phi=0.1$ & .886(.003) & .887(.003) & .760(.005) & .869(.003)  & \bftab.037(.001)   \\ \hline
\end{tabular}

\begin{tabular}{llllll}
\hline
$\pi=0.9,\ k=6$  & \multicolumn{5}{l}{}  \\ 
$\phi=0$ & .700(.005) & .710(.003) & \bftab.000(.000) & .669(.004) & \bftab.000(.000)   \\
$\phi=0.05$ & .777(.003) & .770(.003) & .397(.004) & .738(.003) & \bftab.010(.000)  \\ 
$\phi=0.1$ & .893(.003) & .893(.003) & .774(.005) & .877(.003)  & \bftab.045(.001)   \\ \hline
\end{tabular}

\begin{tabular}{llllll}
\hline
$\pi=0.9, k=10$  & \multicolumn{5}{l}{}  \\ 
$\phi=0$ & .701(.003) & .704(.004) & \bftab.000(.000) & .676(.004) & \bftab.000(.000)   \\
$\phi=0.05$ & .772(.003) & .772(.003) & .384(.004) & .735(.003) & \bftab.010(.001)  \\ 
$\phi=0.1$ & .893(.003) & .891(.003) & .758(.006) & .870(.003)  & \bftab.041(.001)   \\ \hline
\end{tabular}

\caption{NA value proportions for models on the complete test set. The dimension for the simulated matrix is $m=n=100$.}
\label{tab:sim_na_100}
\end{table}
\begin{table}[!htbp]
\centering
\begin{tabular}{llllll}
\hline
$m=n=300$   & CFuser & CFitem & softImpute & blind reg & RNE\\ \hline
$\pi=0.6,\ k=3$  & \multicolumn{5}{l}{}  \\ 
$\phi=0$  & \bftab.000(.000) & \bftab.000(.000) & \bftab.000(.000) & \bftab.000(.000) & \bftab.000(.000)   \\ 
$\phi=0.05$ & .389(.001) & .389(.001) & .389(.001) & .389(.001) & \bftab.010(.000)  \\ 
$\phi=0.1$ & .761(.001) & .761(.001) & .761(.001) & .761(.001) & \bftab.039(.000)  \\  \hline
\end{tabular}

\begin{tabular}{llllll}
\hline
$\pi=0.6,\ k=6$  & \multicolumn{5}{l}{} \\ 
$\phi=0$ & \bftab.000(.000) & \bftab.000(.000) & \bftab.000(.000) & \bftab.000(.000) & \bftab.000(.000)   \\ 
$\phi=0.05$ & .388(.001) & .388(.001) & .388(.001) & .388(.001) & \bftab.010(.000)  \\ 
$\phi=0.1$ & .759(.001) & .759(.001) & .759(.001) & .759(.001) & \bftab.040(.000)  \\  \hline
\end{tabular}

\begin{tabular}{llllll}
\hline
$\pi=0.6, k=10$  & \multicolumn{5}{l}{} \\ 
$\phi=0$ & \bftab.000(.000) & \bftab.000(.000) & \bftab.000(.000) & \bftab.000(.000) & \bftab.000(.000)   \\ 
$\phi=0.05$ & .392(.001) & .392(.001) & .392(.001) & .392(.001)  & \bftab.010(.000)  \\ 
$\phi=0.1$ & .759(.001) & .759(.001) & .759(.001) & .759(.001)  & \bftab.039(.000)  \\  \hline
\end{tabular}

\begin{tabular}{llllll}
\hline
$\pi=0.9,\ k=3$  & \multicolumn{5}{l}{} \\ 
$\phi=0$ & .009(.000) & .008(.000) & \bftab.000(.000) & .001(.000)  & \bftab.000(.000)  \\ 
$\phi=0.05$ & .393(.001) & .393(.001) & .392(.001) & .392(.000)  & \bftab.001(.000)  \\  
$\phi=0.1$ & .759(.002) & .759(.002) & .759(.002) & .759(.000)   & \bftab.041(.000)    \\ \hline
\end{tabular}

\begin{tabular}{llllll}
\hline
$\pi=0.9,\ k=6$  & \multicolumn{5}{l}{} \\ 
$\phi=0$ & .008(.000) & .008(.000) & \bftab.000(.000) & .001(.000) & \bftab.000(.000)  \\ 
$\phi=0.05$ & .389(.001) & .389(.001) & .387(.002) & .387(.002) & \bftab.010(.000)    \\ 
$\phi=0.1$ & .760(.001) & .761(.001) & .760(.001) & .760(.001)  & \bftab.041(.000)    \\ \hline
\end{tabular}

\begin{tabular}{llllll}
\hline
$\pi=0.9, k=10$  & \multicolumn{5}{l}{} \\ 
$\phi=0$ & .007(.000) & .008(.000) & \bftab.000(.000) & .001(.000) & \bftab.000(.000)  \\ 
$\phi=0.05$ & .393(.001) & .393(.001) & .391(.001) & .391(.001) & \bftab.010(.000)    \\ 
$\phi=0.1$ & .760(.001) & .760(.001) & .760(.001) & .760(.001)  & \bftab.040(.000)    \\ \hline
\end{tabular}
\caption{NA value proportions for models on the complete test set. The dimension for the simulated matrix is $m=n=300$.}
\label{tab:sim_na_300}
\end{table}
\begin{table}[!htbp]
\centering
\begin{tabular}{llllll}
\hline
$m=n=100$   & CFuser & CFitem & softImpute & blind reg & RNE\\ \hline
$\pi=0.6,\ k=3$  & \multicolumn{5}{l}{}  \\ 
$\phi=0$ & 1.10(.010) & 1.10(.010) & 1.22(.020) & 1.19(.009) & \bftab 1.07(.002)    \\ 
$\phi=0.05$ & \bftab1.06(.010) & \bftab 1.06(.010) & 1.16(.021) & 1.18(.010) & 1.08(.003)  \\ 
$\phi=0.1$ & \bftab0.97(.010) & 0.97(.011) & 1.06(.019) & 1.09(.009)   & 1.06(.003)  \\ \hline
\end{tabular}

\begin{tabular}{llllll}
\hline
$\pi=0.6,\ k=6$  & \multicolumn{5}{l}{}  \\ 
$\phi=0$ & 1.66(.008) & 1.67(.011) & 3.21(.036) & 1.53(.008) & \bftab1.12(.003)    \\ 
$\phi=0.05$ & 1.63(.010) & 1.62(.011) & 2.92(.034) & 1.48(.009) & \bftab1.12(.003) \\ 
$\phi=0.1$ & 1.62(.018) & 1.61(.019) & 3.00(.059) & 1.47(.014)  & \bftab1.13(.005)    \\ \hline
\end{tabular}

\begin{tabular}{llllll}
\hline
$\pi=0.6, k=10$  & \multicolumn{5}{l}{}   \\ 
$\phi=0$ & 2.25(.015) & 2.24(.017) & 4.95(.071) & 2.05(.015) & \bftab1.21(.004)   \\ 
$\phi=0.05$ & 2.19(.017) & 2.15(.015) & 4.86(.073) & 1.94(.017)  & \bftab1.21(.005)    \\ 
$\phi=0.1$ & 2.12(.019) & 2.12(.025) & 4.68(.082) & 1.89(.020)  & \bftab1.21(.006)    \\ \hline
\end{tabular}

\begin{tabular}{llllll}
\hline
$\pi=0.9,\ k=3$  & \multicolumn{5}{l}{} \\ 
$\phi=0$ & 4.61(.106) & 4.52(.142) & 3.26(.063) & 10.7(.469)  & \bftab1.36(.014)  \\ 
$\phi=0.05$ & 4.40(.145) & 4.30(.101) & 3.14(.062) & 9.16(.271)  & \bftab1.30(.012)    \\ 
$\phi=0.1$ & 4.44(.172) & 3.89(.122) & 3.31(.071) & 7.79(.319)   & \bftab1.29(.015)    \\ \hline
\end{tabular}

\begin{tabular}{llllll}
\hline
$\pi=0.9,\ k=6$  & \multicolumn{5}{l}{} \\ 
$\phi=0$ & 7.68(.181) & 7.48(.162) & 5.91(.100) & 18.4(.589) & \bftab1.59(.013)  \\ 
$\phi=0.05$ & 7.43(.188) & 7.00(.161) & 5.73(.126) & 16.3(.519) & \bftab1.51(.014)    \\ 
$\phi=0.1$ & 6.61(.293) & 7.00(.310) & 6.23(.184) & 14.1(.604)  & \bftab1.43(.024)    \\ \hline
\end{tabular}

\begin{tabular}{llllll}
\hline
$\pi=0.9, k=10$  & \multicolumn{5}{l}{} \\ 
$\phi=0$ & 10.7(.268) & 11.2(.309) & 10.1(.138) & 25.7(.580) & \bftab1.98(.019)  \\ 
$\phi=0.05$ & 11.1(.246) & 9.45(.214) & 10.4(.180) & 26.4(.852) & \bftab1.93(.026)    \\ 
$\phi=0.1$ & 11.0(.604) & 11.4(.520) & 11.5(.359) & 27.4(.997) & \bftab2.02(.051)    \\ \hline
\end{tabular}
\caption{Standardized test error for models on the test set without ``cold-start" entries. The dimension for the simulated matrix is $m=n=100$.}
\label{tab:sim_error_100}
\end{table}
\begin{table}[!htbp]
\centering
\begin{tabular}{llllll}
\hline
$m=n=300$   & CFuser & CFitem & softImpute & blind reg & RNE\\ \hline
$\pi=0.6, k=3$  & \multicolumn{5}{l}{} \\ 
$\phi=0$ & 0.57(.002) & 0.57(.002) & \bftab0.54(.003) & 0.85(.001) & 1.02(.000)  \\ 
$\phi=0.05$ & 0.54(.002) & 0.54(.002) & \bftab0.52(.003) & 0.84(.002) & 1.02(.001)  \\ 
$\phi=0.1$ & 0.51(.003) & 0.51(.003) & \bftab0.51(.004) & 0.82(.002) & 1.02(.001)  \\  \hline
\end{tabular}

\begin{tabular}{llllll}
\hline
$\pi=0.6,\ k=6$  & \multicolumn{5}{l}{} \\ 
$\phi=0$ & \bftab1.02(.002) & \bftab1.02(.002) & 1.23(.006) & 1.13(.002) & 1.04(.001)  \\ 
$\phi=0.05$ & \bftab1.00(.002) & \bftab1.00(.002) & 1.20(.005) & 1.12(.002) & 1.04(.001)  \\ 
$\phi=0.1$ & \bftab1.00(.003) & \bftab1.00(.003) & 1.14(.007) & 1.11(.002) & 1.03(.001)  \\  \hline
\end{tabular}

\begin{tabular}{llllll}
\hline
$\pi=0.6, k=10$  & \multicolumn{5}{l}{} \\ 
$\phi=0$ & 1.32(.003) & 1.32(.002) & 2.09(.009) & 1.26(.002) & \bftab1.07(.001)  \\ 
$\phi=0.05$ & 1.29(.003) & 1.28(.003) & 1.98(.011) & 1.23(.002) & \bftab1.06(.001)  \\ 
$\phi=0.1$ & 1.28(.004) & 1.27(.003) & 1.91(.013) & 1.22(.003) & \bftab1.06(.001)  \\  \hline
\end{tabular}

\begin{tabular}{llllll}
\hline
$\pi=0.9,\ k=3$  & \multicolumn{5}{l}{} \\ 
$\phi=0$ & 1.96(.008) & 1.97(.008) & 1.66(.014) & 2.19(.008) & \bftab1.10(.002)  \\ 
$\phi=0.05$ & 1.73(.010) & 1.77(.009) & 1.62(.016) & 1.91(.011) & \bftab1.10(.003)    \\ 
$\phi=0.1$ & 1.58(.009) & 1.59(.011) & 1.55(.016) & 1.69(.010) & \bftab1.08(.003)    \\ \hline
\end{tabular}

\begin{tabular}{llllll}
\hline
$\pi=0.9,\ k=6$  & \multicolumn{5}{l}{} \\ 
$\phi=0$ & 2.87(.012) & 2.83(.011) & 3.55(.022) & 3.21(.017) & \bftab1.16(.002)  \\ 
$\phi=0.05$ & 2.52(.013) & 2.49(.012) & 3.47(.021) & 2.64(.012) & \bftab1.16(.002)    \\ 
$\phi=0.1$ & 2.21(.014) & 2.21(.016) & 3.54(.041) & 2.29(.018) & \bftab1.15(.003)    \\ \hline
\end{tabular}

\begin{tabular}{llllll}
\hline
$\pi=0.9, k=10$  & \multicolumn{5}{l}{} \\ 
$\phi=0$ & 3.96(.020) & 3.95(.022) & 5.39(.028) & 5.01(.031) & \bftab1.26(.003)  \\ 
$\phi=0.05$ & 3.43(.018) & 3.44(.020) & 5.32(.052) & 4.10(.031) & \bftab1.25(.004)    \\
$\phi=0.1$ & 2.98(.028) & 2.95(.026) & 5.31(.073) & 3.40(.036) & \bftab1.25(.005)    \\ \hline
\end{tabular}
\caption{Standardized test error for models on the test set without ``cold-start" entries. The dimension for the simulated matrix is $m=n=300$.}
\label{tab:sim_error_300}
\end{table}
\begin{table}[!htbp]
\centering
\begin{tabular}{lccc}
\hline
$m=n=100$ & \multicolumn{3}{c}{RNE} \\
            & RMSE    & NA\%   & error   \\ \hline
$\pi=0.6,\ k=3$          &         &      &         \\ 
$\phi=0.05$            & 3.45(.044)       & 0.03(.001)    & 1.14(.007)       \\ 
$\phi=0.1$             & 3.37(.042)       & 0.05(.000)    & 1.15(.007)      \\ \hline \hline
$p=0.6,\ k=6$           &         &      &         \\ 
$\phi=0.05$            & 6.43(.050)       & 0.03(.001)    & 1.23(.007)       \\ 
$\phi=0.1$             & 6.44(.052)       & 0.05(.001)    & 1.22(.006)      \\ \hline \hline
$\pi=0.6, k=10$           &         &      &         \\ 
$\phi=0.05$            & 10.4(.082)       & 0.02(.001)    & 1.45(.012)       \\ 
$\phi=0.1$             & 10.4(.082)       & 0.05(.001)    & 1.40(.008)      \\ \hline \hline
$\pi=0.9,\ k=3$           &         &      &         \\ 
$\phi=0.05$            & 3.62(.047)   & 0.02(.001)    & 1.67(.030)       \\ 
$\phi=0.1$             & 3.58(.045)   & 0.05(.002)    & 1.56(.014)      \\ \hline \hline
$\pi=0.9,\ k=6$           &         &      &         \\ 
$\phi=0.05$            & 6.95(.064)       & 0.03(.001)    & 2.15(.055)       \\ 
$\phi=0.1$             & 6.79(.068)       & 0.06(.002)    & 2.08(.037)      \\ \hline \hline
$\pi=0.9, k=10$           &         &      &         \\ 
$\phi=0.05$            & 11.0(.133)       & 0.03(.002)    & 3.13(.086)       \\ 
$\phi=0.1$             & 10.9(.093)       & 0.05(.001)    & 2.78(.059)      \\ \hline
\end{tabular}

\begin{tabular}{lccc}
\hline
$m=n=300$  & \multicolumn{3}{c}{RNE} \\
               & RMSE    & NA\%   & error   \\ \hline
$\pi=0.6,\ k=3$           &         &      &         \\ 
$\phi=0.05$           & 3.44(.018)       & 0.03(.000)    & 1.04(.001)        \\ 
$\phi=0.1$               & 3.46(.020)       & 0.05(.000)    & 1.04(.001)        \\  \hline \hline
$\pi=0.6,\ k=6$           &         &      &         \\ 
$\phi=0.05$            & 6.50(.032)       & 0.03(.000)    & 1.08(.001)       \\ 
$\phi=0.1$              & 6.40(.032)       & 0.05(.000)    & 1.07(.001)       \\  \hline \hline
$\pi=0.6, k=10$           &         &      &         \\ 
$\phi=0.05$            & 10.5(.039)       & 0.03(.000)    & 1.13(.002)       \\ 
$\phi=0.1$              & 10.5(.043)       & 0.05(.000)    & 1.13(.002)       \\  \hline \hline
$\pi=0.9,\ k=3$           &         &      &         \\ 
$\phi=0.05$             & 3.47(.016)       & 0.03(.000)    & 1.19(.005)       \\  
$\phi=0.1$              & 3.49(.018)       & 0.05(.001)    & 1.17(.003)       \\  \hline \hline
$\pi=0.9,\ k=6$          &         &      &         \\ 
$\phi=0.05$             & 6.61(.036)       & 0.03(.000)    & 1.32(.006)       \\ 
$\phi=0.1$              & 6.55(.029)       & 0.05(.001)    & 1.31(.005)       \\  \hline \hline
$\pi=0.9, k=10$           &         &      &         \\ 
$\phi=0.05$             & 10.7(.044)       & 0.03(.001)    & 1.51(.009)       \\ 
$\phi=0.1$              & 10.7(.042)       & 0.05(.001)    & 1.49(.006)       \\  \hline
\end{tabular}
\caption{RNE performance on test set containing only ``cold-start" cases. NA\% here referred to the proportion of NA values on the cold-start part of the test set specifically.}
\label{tab:sim_coldstart}
\vspace{-60pt} 
\end{table}

\subsection{Empirical study}

We evaluate the performance of the proposed algorithm by using several commonly experimented datasets: MoveieLens 0.1M dataset \citep{harper2015movielens}, Jester dataset 4 \citep{goldberg2001eigentaste}, Last.fm dataset \citep{Bertin-Mahieux2011}, and steam video games (\href{https://www.kaggle.com/datasets/tamber/steam-video-games}{https://www.kaggle.com/datasets/tamber/steam-video-games}). MovieLens 0.1M contains 100,836 tuples defined by $\texttt{[user, movie, rating]}$ for 610 users IDs and 9,724 movies IDs, and the ratings are discrete as $[0.5,~1,~1.5,~2,~2.5,~3,~3.5,~4,~4.5,~5]$. Jester dataset 4 contains 106,489 tuples defined by $\texttt{[user, joke, rating]}$ for 7,699 user IDs and 136 joke IDs after filtering out those with 99 (null) as ratings. The ratings range from $-10$ to $10$. Last.fm contains 92,834 tuples defined by $\texttt{[user, artist, listeningCount]}$ for 1,892 user IDs and 17,632 artist IDs. Steam video games contains 70,477 tuples defined by $\texttt{[user, game, playingTime]}$ for 11,350 user IDs and 3,600 game IDs after accumulating the hours for one user if his or her behaviour is labeled as $\texttt{play}$. Log transformations were made to the time of play in the steam video games and the time of listening in the Last.fm. 

For each dataset, a random sub-sample of 30\% of users and items was extracted for each repetition. The subset was then divided into training and test sets, and we evaluated all methods separately at the training scale (0.9, 0.7, 0.5) and the test scale (0.1, 0.3, 0.5). The repetition round was set as 100. We chose RMSE and the NA value proportions as the metrics for evaluation and output the mean and standard error for metrics. The results were again visualized on the basis of splitting the test set into a cold-start part and a non-cold-start part. The details of parameter and model choosing can be found in the previous section of simulation.

\begin{figure}[!htbp]
\centering
\includegraphics[width=0.9\textwidth]{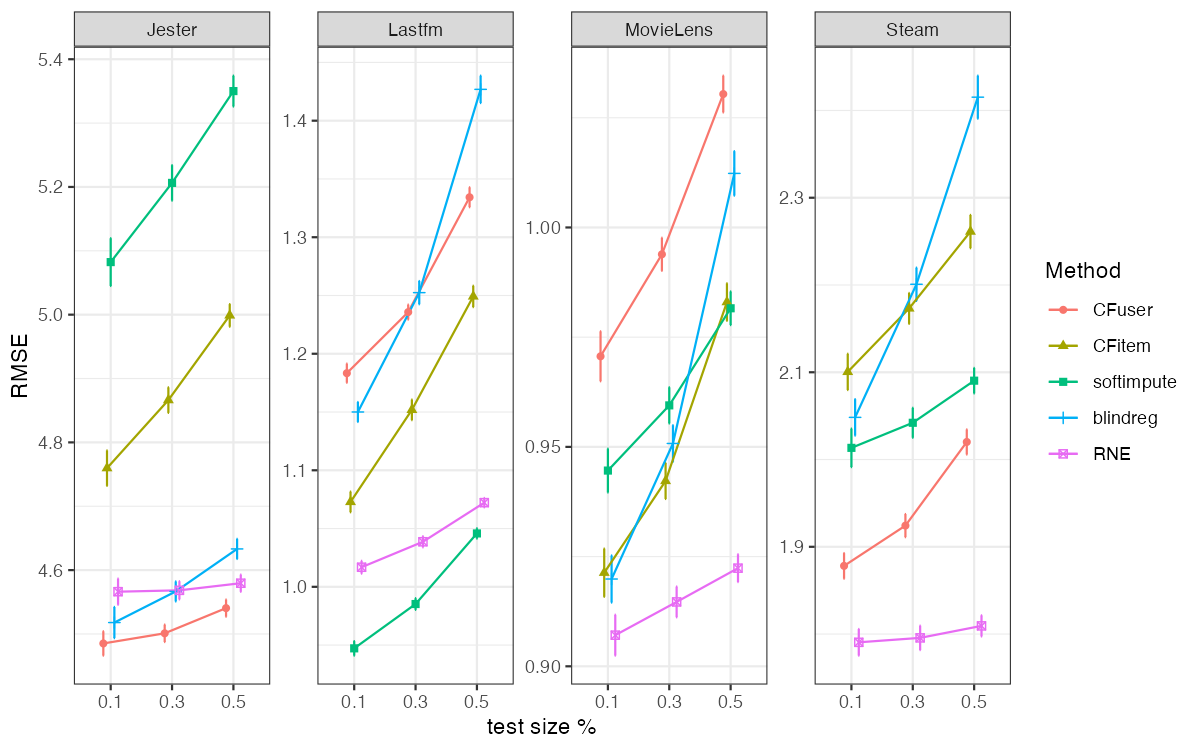}
\includegraphics[width=0.9\textwidth]{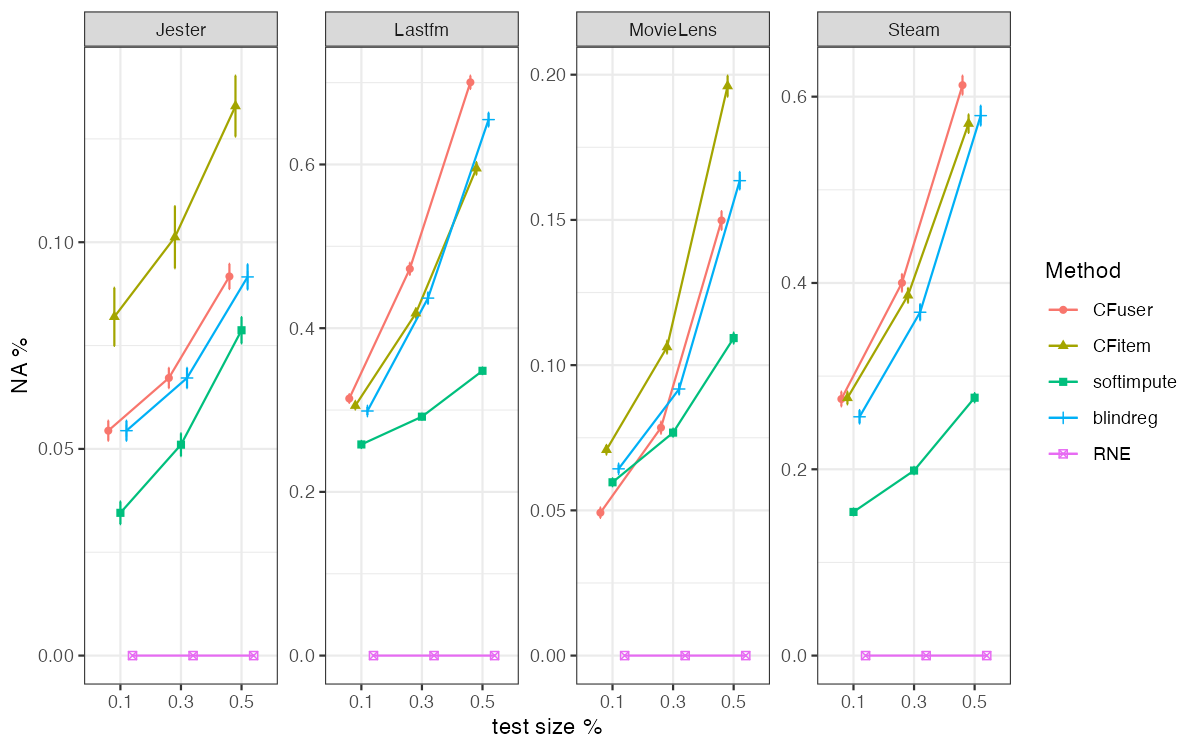}
\caption{Performance on non-cold-start part. NA\% referred to the proportions of NA values in the test part excluding the cold-start part.}
\label{fig:noncold}
\end{figure}

\begin{figure}[!htbp]
\centering
\includegraphics[width=0.9\textwidth]{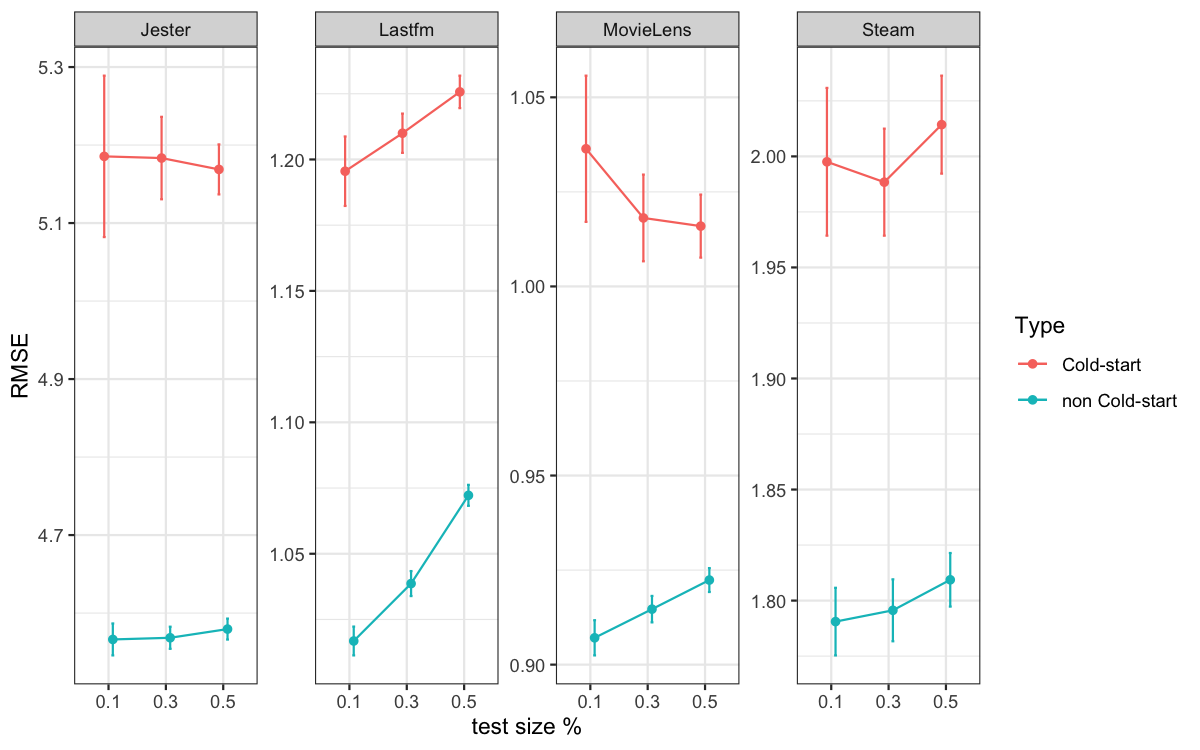}
\includegraphics[width=0.9\textwidth]{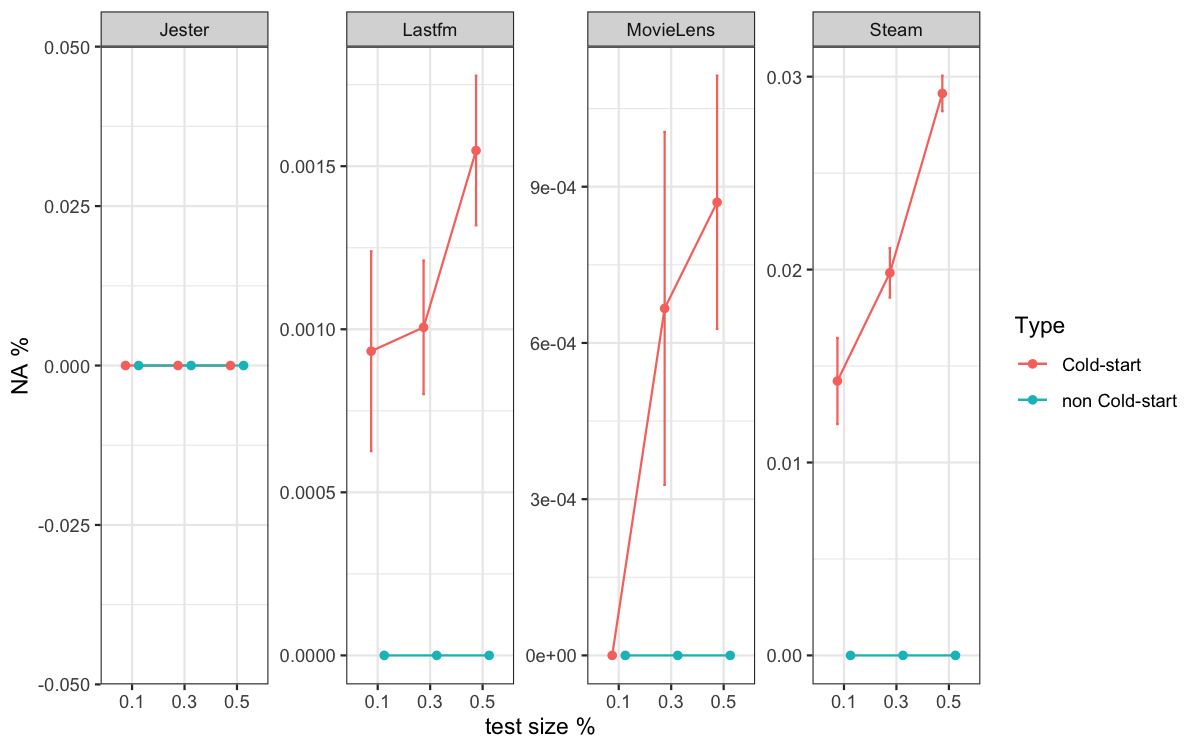}
\caption{Performance on cold-start part for the proposed methods.}
\label{fig:cold}
\end{figure}

Figure~\ref{fig:noncold} showed RMSE and NA proportion comparisons for user-based collaborative filtering, item-based collaborative filtering, soft impute, blind regression, RNE on the test set excluding ``cold-start" part. Both the curves for RMSE and NA proportions exhibited increasing patterns along the increasing test size proportion. This was reasonable as the smaller training samples the less information to pool for prediction. It could be noted that RNE outperformed all other models in the MovieLens and Steam datasets, was close to soft impute in the Lastfm dataset, was close to blind regression and slightly inferior to user-based collaborative filtering in the Jester dataset. The fluctuations might be caused by the nature of datasets while we could claim that the performance of RNE were generally satisfying. When looking at NA proportion, RNE indisputably performed better than all other methods because they produced the least values.

Figure~\ref{fig:cold} compared RMSE and NA proportion produced by RNE for the testing set without ``cold-start" cases and testing set with only ``cold-start" cases. The dashed lines were the same as the corresponding ones in Figure~\ref{fig:cold} and they served benchmark to identify the difference introduced by ``cold-start" cases, and the solid lines represented the results from testing set with ``cold-start" only. We found that RMSE were deteriorated while NA proportions changed within rather limited ranges. Both of the results validated that the generalizability and scalability to real world cases for RNE.

\section{Discussion}
This article presents a smoothing recommender system RSE based on the row- and column-wise $L_2$ norm to improve prediction accuracy with the observed rating matrix only. It provides a flexible framework for integrating and exploiting latent features based on user and item data and relaxes the definition of neighbors and distance computation so that the effective sample size is increased for higher prediction accuracy and less failures due to missing information, i.e., ``cold-start" problem.

There are several future directions based on the proposed method. We can further consider the construction of kernel functions within the estimator and make use of the independence between user-related features and item-related features to propose a weighted average of two univariate kernels, while this estimator would require more discussions on the rationality and theoretical support. Second, we simplify the missing mechanism into missing completely as random while many researches have pointed out that bias introduced by varying missing rates or mechanism such as missing at random should be considered in the modeling \citep{marlin2009collaborative, ying2006leveraging, chen2020bias}. Besides, our model takes into consideration the observed rating matrix only while external covariate information such as demographic information, social network, and geo-spatial relationships, can contribute to the prediction as well \citep{mao2019matrix, dai2019smooth}.

\newpage

\vskip 0.2in
\bibliography{sample}

\newpage

\section{Appendix}\label{appendix}

\subsection{Proof of Proposition~\ref{prop2}}

WLOG, we choose user-wise distance for example and proof of item-wise distance can be generalized. We first expand equation \ref{dhat} as
\bea
    \widehat{d}_{u,v}^2 &=&\frac{1}{m \pi^2}
    \sum_{i \in \mathcal{I}_{uv}} (a_{u, i} - a_{v, i})^2 + (\frac{1}{|\mathcal{I}_{uv}|} - \frac{1}{m \pi^2})\sum_{i \in \mathcal{I}_{uv}} (a_{u, i} - a_{v, i})^2 \nn \\
    &=&\frac{1}{m \pi^2}\sum_{i \in \mathcal{I}_{uv}} (a_{u, i} - a_{v, i})^2 + O_p(m^{-1/2} \pi^{-1})\nn
\eea
by using $|\mathcal{I}_{uv}| - m\pi^2 = O_p(\sqrt{m} \pi)$ derived from  the definition of $\mathcal{I}_{uv}$ and Assumption~\ref{asu1}.
\bea
    \mathbb{E}(\widehat{d}_{u,v}^2 \mid\bA_{(u)}, \bA_{(v)})&=&\frac{p^2\sum_{t=1}^m(a_{u,t}-a_{v,t})^2}{m\pi^2}=\frac{1}{m}\|\bA_{(u)}-\bA_{(v)}\|_2^2  \nn \\
    \mathrm{Var}(\widehat{d}_{u,v}^2 \mid\bA_{(u)}, \bA_{(v)})&=& \frac{\sum_{t=1}^m(a_{u,t}-a_{v,t})^4\mathrm{Var}(\delta_{u,t}\delta_{v,t})}{m^2\pi^4} \nn \\
    &=&
    \frac{\sum_{t=1}^m(a_{u,t}-a_{v,t})^4\mathrm{Var}[\mathbb{E}(\delta_{u,t}\delta_{v,t})\mid \delta_{v,t}] + \mathbb{E}[\mathrm{Var}(\delta_{u,t}\delta_{v,t})\mid \delta_{v,t})]}{m^2\pi^4} \nn \\
    &=&
    \frac{\sum_{t=1}^m(a_{u,t}-a_{v,t})^4[\mathrm{Var}(\delta_{u,t})\mathbb{E}^2(\delta_{u,t}) +\mathrm{Var}(\delta_{u,t})\mathrm{Var}(\delta_{v,t})+\mathrm{Var}(\delta_{v,t})\mathbb{E}^2(\delta_{v,t})]}{m^2\pi^4} \nn \\
    &=& \frac{\pi^2(1-\pi)^2\sum_{t=1}^m(a_{u,t}-a_{v,t})^4}{m^2\pi^4} \nn \\
    &=& O_p(m^{-1}\pi^{-1}) \nn
\eea
so that 
\bea
    \mathbb{E}(\widehat{d}_{u,v}^2-m^{-1}\|\bA_{(u)}-\bA_{(v)}\|_2^2\mid\bA_{(u)}, \bA_{(v)})
    &=& 0 \nn \\
    \mathrm{Var}(\widehat{d}_{u,v}^2-m^{-1}\|\bA_{(u)}-\bA_{(v)}\|_2^2\mid\bA_{(u)}, \bA_{(v)})
    &=&
    \mathbb{E}\big\{\mathrm{Var}(\widehat{d}_{u,v}^2-m^{-1}\|\bA_{(u)}-\bA_{(v)}\|_2^2\mid \bA_{(u)}, \bA_{(v)})\big\} \nn \\
    && +
    \mathrm{Var}\big\{\mathbb{E}(\widehat{d}_{u,v}^2-m^{-1}\|\bA_{(u)}-\bA_{(v)}\|_2^2\mid \bA_{(u)}, \bA_{(v)})\big\} \nn \\
    &=& O_p(m^{-1/2}\pi^{-1}) \nn
\eea
thus $ \big | \widehat{d}_{u,v}^2 - m^{-1}\|\bA_{(u)}-\bA_{(v)}\|_2^2 \big | = O_p(m^{-1/2}\pi^{-1})$.

\subsection{Proof of Proposition~\ref{prop3}}
WLOG, we choose user-wise distance for example and proof of item-wise distance can be generalized. By the assumption that $|\widehat{\sigma}-\sigma| = O_p(m^{-1/2}n^{-1/2}\pi^{-1})$,
\bea
    m^{-1}\|\bA_{(u)}-\bA_{(v)}\|_2^2 - 2\widehat{\sigma}^2
    =
    \frac{1}{m}\sum_{t=1}^m\left[(z_{u,t}-z_{v,t})^2+(\epsilon_{u,t}-\epsilon_{v,t})^2-2(\epsilon_{u,t}-\epsilon_{v,t})(z_{u,t}-z_{u,t})\right] - 2\widehat{\sigma}^2 \nn
\eea
\bea    
    \mathbb{E} \big( m^{-1} \|\bA_{(u)}-\bA_{(v)}\|_2^2 - 2\widehat{\sigma}^2 \mid \bZ_{(u)}, \bZ_{(v)} \big)
    &=&
    m^{-1}\|\bZ_{(u)}-\bZ_{(v)}\|_2^2 + \mathbb{E}\big(2\sigma^2 - 2\widehat{\sigma}^2\big) \nn \\
    &=&
    m^{-1}\|\bZ_{(u)}-\bZ_{(v)}\|_2^2 \nn + O_p(m^{-1/2}n^{-1/2}\pi^{-1})\\
    \mathrm{Var} \big( m^{-1}\|\bA_{(u)}-\bA_{(v)}\|_2^2 - 2\widehat{\sigma}^2 \mid \bZ_{(u)}, \bZ_{(v)} \big)
    &=&
    m^{-2}\sum_{t=1}^m[\mathrm{Var}(z_{u,t}-z_{v,t})^2+
                \mathrm{Var}(\epsilon_{u,t}-\epsilon_{v,t})^2 \nn \\
    && + 4\mathrm{Var}\{(\epsilon_{u,t}-\epsilon_{v,t})(z_{u,t}-z_{u,t})\} \nn \\
    && -4\mathrm{Cov}\{(z_{u,t}-z_{v,t})^2, (\epsilon_{u,t}-\epsilon_{v,t})(z_{u,t}-z_{u,t})\} \nn \\
    && -4\mathrm{Cov}\{(\epsilon_{u,t}-\epsilon_{v,t})^2, (\epsilon_{u,t}-\epsilon_{v,t})(z_{u,t}-z_{u,t})\}] \nn \\
    &=& m^{-2}\sum_{t=1}^m [\mathrm{Var}(\epsilon_{u,t}-\epsilon_{v,t})^2 + 4\mathrm{Var}\{(\epsilon_{u,t}-\epsilon_{v,t})(z_{u,t}-z_{u,t})\}] \nn 
\eea
as $z_{u,t}$ and $z_{v,t}$ are all fixed and $\epsilon_{u,t}$ and $\epsilon_{v,t}$ are independent random error terms. Further,
\bea
    \mathrm{Var}(\epsilon_{u,t}-\epsilon_{v,t})^2
    &=&
    \mathrm{Var}(\epsilon_{u,t}^2+\epsilon_{v,t}^2-2\epsilon_{u,t}\epsilon_{v,t}) \nn \\
    &=&
    \mathrm{Var}\epsilon_{u,t}^2+\mathrm{Var}\epsilon_{v,t}^2+4\mathrm{Var}\epsilon_{u,t}\epsilon_{v,t}-4\mathrm{Cov}(\epsilon_{u,t}^2, \epsilon_{u,t}\epsilon_{v,t})-4\mathrm{Cov}(\epsilon_{v,t}^2, \epsilon_{u,t}\epsilon_{v,t}) \nn \\
    &=&
    2\sigma^4+2\sigma^4+4\sigma^4 \nn \\
    &=&
    8\sigma^4 \nn \\
    4\mathrm{Var}\{(\epsilon_{u,t}-\epsilon_{v,t})(z_{u,t}-z_{u,t})\}
    &=&
    4(z_{u,t}-z_{v,t})^2\cdot2\sigma^2=8\sigma^2(z_{u,t}-z_{v,t})^2 \nn
\eea
Overall,
\bea
    \mathrm{Var} \big( m^{-1}\|\bA_{(u)}-\bA_{(v)}\|_2^2 - 2\widehat{\sigma}^2 \mid \bZ_{(u)}, \bZ_{(v)} \big)
    &=&
    \frac{1}{m^2}\{8m\sigma^4+8\sigma^2\sum_{t=1}^m(z_{u,t}-z_{v,t})^2\} = O_p(m^{-1}) \nn
\eea
then similar to Proposition~\ref{prop2}, we can get $\big| m^{-1}\|\bA_{(u)}-\bA_{(v)}\|_2^2 - 2\widehat{\sigma}^2 - m^{-1}\|\bZ_{u,\cdot}-\bZ_{v,\cdot}\|_2^2 \big| = O_p(m^{-1/2}n^{-1/2}\pi^{-1})$.

\subsection{Proof of Proposition~\ref{prop4}}
WLOG, we choose user-wise distance for example and proof of item-wise distance can be generalized.
\bea
    \|\bZ_{(u)} - \bZ_{(v)}\|_2^2
    &=&
    \|(x_{u,1}-x_{v,1})\boldsymbol{q}_1^{\T}+\cdots+(x_{u,k}-x_{v,k})\boldsymbol{q}_k^{\T} \|_2^2 \nn \\
    &=&
    \{(x_{u,1}-x_{v,1})\boldsymbol{q}_1^{\T}+\cdots+(x_{u,k}-x_{v,k})\boldsymbol{q}_k^{\T}\}
    \{(x_{u,1}-x_{v,1})\boldsymbol{q}_1+\cdots+(x_{u,k}-x_{v,k})\boldsymbol{q}_k\}  \nn \\
    &=&
    (x_{u,1}-x_{v,1})^2\cdots+(x_{u,k}-x_{v,k})^2 \nn \\
    &=&
    \|\boldsymbol{x}_u-\boldsymbol{x}_v\|_2^2 \nn
\eea

\subsection{Proof of Lemma~\ref{lemma1}}
\bea
    LHS &=& |g(\bx_u, \by_i) - g(\bx_u, \by_j) +g(\bx_u, \by_j) - g(\bx_v, \by_j)| \nn \\
    &\leqslant& |g(\bx_u, \by_i) - g(\bx_u, \by_j)| + |g(\bx_u, \by_j) - g(\bx_v, \by_j)| \nn \\
    &=& |\frac{x_{u,1}}{d_1}(y_{1,i}-y_{1,j})+\cdots+\frac{x_{u,k}}{d_k}(y_{k,i}-y_{k,j})| + |\frac{y_{1,j}}{d_1}(x_{u,1}-x_{v,1})+\cdots+\frac{y_{k,j}}{d_k}(x_{u,k}-x_{v,k})| \nn \\
    &\leqslant& \sqrt{k} \big\{p_{u, 1}^2(y_{1,i}-y_{1,j})^2 + \cdots + p_{u, k}^2(y_{k,i}-y_{k,j})^2\big\}^{1/2} \nn \\
    & & + \sqrt{k} \big\{q_{1,j}^2(x_{u,1}-x_{v,1})^2+\cdots+q_{k,j}^2(x_{u,k}-x_{v,k})^2\big\}^{1/2} \nn \\
    &\leqslant&  \sqrt{k}\|\bfp_{(u)}\|_{\max}\|\bx_u-\bx_v\|_2 + \sqrt{k}\|\bfq_{j}\|_{\max} \|\by_i-\by_j\|_2 \nn
\eea
where $\|\cdot\|_{\max}$ denotes the maximum norm which is the maximum value within the given vector. According to Assumption~\ref{asu3} and \ref{asu4}, we can derive $L=2(\sqrt{mk/n}+\sqrt{nk/m})=4\sqrt{k}$.

\subsection{Proof of Lemma~\ref{lemma2}}
For a given target pair $(u,i)$, we can alternatively define its corresponding radial neighbor set as
\bea
\mathcal{N}_{ui}^{\mathrm{Rdl}} &:=& \{(v,j): \delta_{v,j}=1, v\in\mathcal{U}_{ij}, j\in[m], |\mathcal{U}_{ij}|\geqslant \beta \} \nn \\
&& \cup
\{(v',j'): \delta_{v',j'}=1, v'\in[n], j'\in\mathcal{I}_{uv'}, |\mathcal{I}_{uv'}|\geqslant\beta \} \label{radial_neighbor2}
\eea

The derivation from definition~(\ref{radial_neighbor1}) to (\ref{radial_neighbor2}) is obvious as $\mathcal{I}_{uv}$ and $\mathcal{U}_{ij}$ contain the common observed entries between rows and columns so that $\widehat{d}_{u,v}$ and $\widehat{d}_{i,j}$ exist. 

Given a specific $(u,i)$, the probability of observing ratings for $i$ and any $j\in[m]$ for the same user is $\mathbb{P}(\delta_{u,i}=1)\mathbb{P}(\delta_{u,j}=1)=\pi^2$ by Assumption~\ref{asu1}. Therefore, it follows that $|\mathcal{U}_{ij}| \sim Bin(n, \pi^2)$. By Chernoff bound, for $\alpha \in (0,1)$, we have
\bea
    \mathbb{P}(|\mathcal{U}_{ij}| \notin (1 \pm \alpha)n\pi^2) &\leqslant&
    2\exp \bigg(-\frac{\alpha^2}{3}n\pi^2\bigg) \nn
\eea
so that we can claim $|\mathcal{U}_{ij}|$ concentrates around $n\pi^2$. Similarly, we claim $|\mathcal{I}_{uv'}|$ concentrates around $m\pi^2$ and have
\bea
    \mathbb{P}(|\mathcal{I}_{uv'}| \notin (1 \pm \alpha)m\pi^2) &\leqslant&
    2\exp \bigg(-\frac{\alpha^2}{3}m\pi^2\bigg) \nn
\eea
Due to duplicated counting, we know that $|\mathcal{N}_{ui}^{\mathrm{Rdl}}|\leqslant m|\mathcal{U}_{ij}|+n|\mathcal{I}_{uv'}|$ for $j\in[m]$ and $v'\in[n]$. Apply union bound and we can get
\bea
    \mathbb{P}\big( |\mathcal{N}_{ui}^{\mathrm{Rdl}}| \notin (1 \pm \alpha)mn\pi^2\big) 
    &\leqslant&
    2m\exp \bigg(-\frac{\alpha^2}{3}n\pi^2\bigg) \wedge 2n\exp \bigg(-\frac{\alpha^2}{3}m\pi^2\bigg) \nn
\eea
so that $C_1mnp^2 \leqslant |\mathcal{N}_{ui}^{\mathrm{Rdl}}| \leqslant C_2mnp^2$ where $C_1$ and $C_2$ are constants.

\subsection{Proof of Lemma~\ref{lemma3}}
Given a pair of $(u,i)$, we let $r_v:=\|\bx_v-\bx_u\|_2$ and $s_j:=\|\by_j-\by_i\|_2$ by the simplicity of notations. We use $f^*_{u,i}$ to denote the density function for $(r_v, s_j)$.
\bea
\mathbb{E}\{\widetilde{f}(\bx_{u},\by_i)\}
    &=& 
\iint
\frac{1}{h_1h_2}
K\bigg(\frac{r_{v}}{h_1},
        \frac{s_j}{h_2}\bigg)f_{u,i}^*(r_{v}, s_j)dr_{v}ds_{j} \nn \\
    &=&
\iint K(a,b)f_{u,i}^*(ah_1, bh_2)dadb \nn \\
\mathrm{Var}\bigg\{K\bigg(\frac{r_v}{h_1}, \frac{s_j}{h_2}\bigg)\bigg\}
&=& \mathbb{E}\bigg\{K^2\bigg(\frac{r_v}{h_1}, \frac{s_j}{h_2}\bigg)\bigg\}-
\mathbb{E}^2\bigg\{K\bigg(\frac{r_v}{h_1}, \frac{s_j}{h_2}\bigg)\bigg\} \nn \\
&=&
\iint K^2\bigg(\frac{r_v}{h_1}, \frac{s_j}{h_2}\bigg)f_{u,i}^*(r_v, s_j)dr_vds_j 
- h_1^2h_2^2\mathbb{E}^2\{\widetilde{f}(\bx_{u},\by_i)\} \nn \\
&=&
h_1h_2\iint K^2(a, b)f_{u,i}^*(ah_1, bh_2)dadb
- h_1^2h_2^2\mathbb{E}^2\{\widetilde{f}(\bx_{u},\by_i)\} \nn
\eea
\bea
\mathrm{Cov}\bigg\{K\bigg(\frac{r_v}{h_1}, \frac{s_j}{h_2}\bigg), 
                K\bigg(\frac{r_{v'}}{h_1}, \frac{s_j}{h_2}\bigg)\bigg\}
&=& \mathbb{E}\bigg\{K\bigg(\frac{r_v}{h_1}, \frac{s_j}{h_2}\bigg) K\bigg(\frac{r_{v'}}{h_1}, \frac{s_j}{h_2}\bigg)\bigg\} 
- \mathbb{E}\bigg\{K\bigg(\frac{r_v}{h_1}, \frac{s_j}{h_2}\bigg)\bigg\}
\mathbb{E}\bigg\{K\bigg(\frac{r_{v'}}{h_1}, \frac{s_j}{h_2}\bigg)\bigg\} \nn \\
&=&
\iiint K\bigg(\frac{r_v}{h_1}, \frac{s_j}{h_2}\bigg) K\bigg(\frac{r_{v'}}{h_1}, \frac{s_j}{h_2}\bigg)f_{u,i}^*(r_v, s_j)f_{u,i}^*(r_{v'}, s_j)dr_v dr_{v'} ds_j \nn \\
&& - h_1^2h_2^2\mathbb{E}^2\{\widetilde{f}(\bx_u, \by_i)\} \nn \\
&=&
h_1^2h_2\iiint K(a, b)K(a', b)f_{u,i}^*(ah_1, bh_2)f_{u,i}^*(a'h_1, bh_2)da da' db \nn \\
&& - h_1^2h_2^2\mathbb{E}^2\{\widetilde{f}(\bx_u, \by_i)\} \nn \\
\mathrm{Var}\{\widetilde{f}(\bx_{u},\by_i)\}
&=& \frac{1}{h_1h_2|\mathcal{N}_{ui}^{\mathrm{Rdl}}|}\iint K^2(a, b)f_{u,i}^*(ah_1, bh_2)dadb \nn \\
&& + \frac{\sum_j|\mathcal{N}^1_{ui,j}|}{h_2|\mathcal{N}_{ui}^{\mathrm{Rdl}}|^2} \iiint K(a, b)K(a', b)f_{u,i}^*(ah_1, bh_2)f_{u,i}^*(a'h_1, bh_2)da da' db \nn \\
&& + \frac{\sum_v|\mathcal{N}^2_{ui,v}|}{h_1|\mathcal{N}_{ui}^{\mathrm{Rdl}}|^2} \iiint K(a, b)K(a, b')f_{u,i}^*(ah_1, bh_2)f_{u,i}^*(ah_1, b'h_2)da da db' \nn \\
&& - \frac{|\mathcal{N}_{ui}^{\mathrm{Rdl}}|+\sum_{j}|\mathcal{N}^1_{ui,j}|+\sum_{v}|\mathcal{N}^2_{ui,v}|}{|\mathcal{N}_{ui}^{\mathrm{Rdl}}|^2}\mathbb{E}^2\{\widetilde{f}(\bx_u, \by_i)\} \nn
\eea 
We here define subset of radial neighbors. Given a pair of $(u,i)$,
\bea
\mathcal{N}_{ui,j}^1 
&:=&
\{(a^1, b^1) \in \mathcal{U}_{ij}\times \{j\}\ s.t.\ \delta_{a^1,b^1}=1, j\in[m]\} \nn \\
\mathcal{N}_{ui,v}^2 
&:=&
\{(a^2, b^2) \in \{v\} \times \mathcal{I}_{uv}\ s.t.\ \delta_{a^2,b^2}=1, v\in[n]\} \nn
\eea
so that $\mathcal{N}^1_{ui,j}$ denote the subset where two different users have rated one common item and $\mathcal{N}^2_{ui,v}$ denote the subset where each user has rated two different items. Both of them actually hold the correlated neighbors so that
$\sum_{j}|\mathcal{N}_{ui,j}^1|=|\mathcal{N}_{ui}^{\mathrm{Rdl}}|$, $\sum_{v}|\mathcal{N}_{ui,v}^2|=|\mathcal{N}_{ui}^{\mathrm{Rdl}}|$, and we know that $C_1mn\pi^2 \leqslant |\mathcal{N}_{ui}^{\mathrm{Rdl}}| \leqslant C_2mn\pi^2$. We claim that $\frac{\widetilde{f}(\bx_u, \by_i)}{\mathbb{E}\{\widetilde{f}(\bx_u, \by_i)\}}\stackrel{p}{\rightarrow} 1$ by showing
\bea
\mathbb{E}\bigg[ \frac{\widetilde{f}(\bx_u, \by_i)}{\mathbb{E}\{\widetilde{f}(\bx_u, \by_i)\}} \bigg] &=& 1 \nn \\
\mathrm{Var}\bigg[ \frac{\widetilde{f}(\bx_u, \by_i)}{\mathbb{E}\{\widetilde{f}(\bx_u, \by_i)\}} \bigg] 
&=& \frac{\mathrm{Var}\{\widetilde{f}(\bx_u, \by_i)\}}{\mathbb{E}^2\{\widetilde{f}(\bx_u, \by_i)\}} = O_p(m^{-1}n^{-1}\pi^{-2}) \nn
\eea
as $\pi=O_p(m^{-1/2}n^{-1/2})$ and $h_1,h_2=O_p(m^{-1/2}n^{-1/2}\pi^{-1})$. General conditions for the kernel function are required including $K$ is a symmetric function, $\iint K(a,b) = 1$, $\iint K^2(a,b)\leqslant \infty$, $\iiint K(a,b)K(a',b) \leqslant \infty$, $\iiiint K(a,b)K(a',b') \leqslant\infty$. All the conditions are assumed to be satisfied for the subsequent proofs.

\subsection{Proof of Theorem~\ref{theorem1}}
The proposed estimator is
\bea
\widehat{z}_{u,i}=\widehat{g}(\bx_{u}, \by_i)=
    \sum_{(v, j) \in \mathcal{N}_{ui}^{\mathrm {Rdl }}}
        \frac{K\big(\frac{\|\bx_{v} - \bx_{u}\|_2}{h_1}, 
                \frac{\|\by_{i} - \by_{j}\|_2}{h_2}\big)}
        {\sum_{(v, j) \in \mathcal{N}_{ui}^{\mathrm {Rdl }}} 
              K\big(\frac{\|\bx_{v} - \bx_{u}\|_2}{h_1}, 
                \frac{\|\by_{i} - \by_{j}\|_2}{h_2)}\big) } z_{v,j}, \nn
\eea
and by definitions we have
\bea
    z_{u,i} = g(\bx_{u}, \by_i)+\epsilon_{u,i}, \nn
\eea
where $\mathbb{E}(\epsilon_{u,i}|\bx_{u},\by_i)=0$. Expand $z_{v,j}$ in the neighborhood of pair $(u,i)$ so that
\bea
    z_{v,j} =g(\bx_{v}, \by_j)+\epsilon_{v,j}
=g(\bx_{u}, \by_i)+\{g(\bx_{v}, \by_j)-g(\bx_{u}, \by_i)\}+\epsilon_{v,j} \nn
\eea
and therefore
\bea
&&\frac{1}{h_1h_2|\mathcal{N}_{ui}^{\mathrm{rdl}}|}
    \sum_{(v,j)\in\mathcal{N}_{ui}^{\mathrm{rdl}}} K\bigg(\frac{\|\bx_v-\bx_u\|_2}{h_1}, 
            \frac{\|\by_j-\by_i\|_2}{h_2}
    \bigg) z_{v,j} \nn \\
&=& \frac{1}{h_1h_2|\mathcal{N}_{ui}^{\mathrm{rdl}}|} \sum_{(v,j)\in\mathcal{N}_{ui}^{\mathrm{rdl}}}
    K\bigg(\frac{\|\bx_v-\bx_u\|_2}{h_1}, 
            \frac{\|\by_j-\by_i\|_2}{h_2}
    \bigg) g(\bx_u, \by_i) \nn \\
&&+\frac{1}{h_1h_2|\mathcal{N}_{ui}^{\mathrm{rdl}}|} \sum_{(v,j)\in\mathcal{N}_{ui}^{\mathrm{rdl}}}
    K\bigg(\frac{\|\bx_v-\bx_u\|_2}{h_1}, 
            \frac{\|\by_j-\by_i\|_2}{h_2}
    \bigg)
    \{g(\bx_{v}, \by_j)-g(\bx_{u}, \by_i)\} \nn \\
&&+\frac{1}{h_1h_2|\mathcal{N}_{ui}^{\mathrm{rdl}}|} \sum_{(v,j)\in\mathcal{N}_{ui}^{\mathrm{rdl}}}
    K\bigg(\frac{\|\bx_v-\bx_u\|_2}{h_1}, 
            \frac{\|\by_j-\by_i\|_2}{h_2}
    \bigg) \epsilon_{v,j} \nn \\
&=& \widetilde{f}(\bx_{u},\by_i) g(\bx_{u},\by_i)+\widehat{m}_1(\bx_{u},\by_i)+\widehat{m}_2(\bx_{u},\by_i) \label{thm1_decomposition} \\
\widehat{m}_1(\bx_{u},\by_i) 
&:=& 
\frac{1}{h_1h_2|\mathcal{N}_{ui}^{\mathrm{rdl}}|} \sum_{(v,j)\in\mathcal{N}_{ui}^{\mathrm{rdl}}}
    K\bigg(\frac{\|\bx_v-\bx_u\|_2}{h_1}, 
            \frac{\|\by_j-\by_i\|_2}{h_2}
    \bigg)
    \{g(\bx_{v}, \by_j)-g(\bx_{u}, \by_i)\} \nn \\
\widehat{m}_2(\bx_{u},\by_i) 
&:=&
\frac{1}{h_1h_2|\mathcal{N}_{ui}^{\mathrm{rdl}}|} \sum_{(v,j)\in\mathcal{N}_{ui}^{\mathrm{rdl}}}
    K\bigg(\frac{\|\bx_v-\bx_u\|_2}{h_1}, 
            \frac{\|\by_j-\by_i\|_2}{h_2}
    \bigg) \epsilon_{v,j} \nn
\eea
Divide both sides of the equation (\ref{thm1_decomposition}) by $\widetilde{f}(\bx_{u},\by_i)$, then
\bea
    \widehat{z}_{u,i}
    &=&
    \widehat{g}(\bx_{u},\by_i)
    =
    g(\bx_{u},\by_i)+
    \frac{\widehat{m}_1(\bx_{u},\by_i)}{\widetilde{f}(\bx_{u},\by_i)}+
    \frac{\widehat{m}_2(\bx_{u},\by_i)}{\widetilde{f}(\bx_{u},\by_i)} \nn \\
    &=& z_{u,i} +
    \frac{\widehat{m}_1(\bx_{u},\by_i)}{\widetilde{f}(\bx_{u},\by_i)}+
    \frac{\widehat{m}_2(\bx_{u},\by_i)}{\widetilde{f}(\bx_{u},\by_i)} \label{expectation}
\eea
We prove the consistency of $\widehat{z}_{u,i}$ to $z_{u,i}$ by showing $\frac{\widehat{m}_1(\bx_{u},\by_i)}{\widetilde{f}(\bx_{u},\by_i)} \to 0$ and $\frac{\widehat{m}_2(\bx_{u},\by_i)}{\widetilde{f}(\bx_{u},\by_i)} \to 0$. First we take $\frac{\widehat{m}_2(\bx_{u},\by_i)}{\widetilde{f}(\bx_{u},\by_i)}$. By $\mathbb{E}(\epsilon_{v,j} \mid \bx_{u}, \by_i)=0$, it follows that
\bea
\mathbb{E}\bigg\{K\bigg(\frac{\|\bx_{u}-\bx_{v}\|_2}{h_1}, \frac{\|\by_i-\by_j\|_2}{h_2}\bigg)\cdot \epsilon_{v,j}\bigg\}&=&0 \nn
\eea
and thus $\mathbb{E}\{\widehat{m}_2(\bx_{u}, \by_i)\}=0$. By Lemma~\ref{lemma3},
\bea
    \mathbb{E}\bigg\{\frac{\widehat{m}_2(\bx_{u},\by_i)}{\widetilde{f}(\bx_{u},\by_i)}\bigg\} 
    =
    \frac{\mathbb{E}\{ \widehat{m}_2(\bx_{u},\by_i) \}}
    {\mathbb{E}\{\widetilde{f}(\bx_{u},\by_i)\}} = 0 \nn
\eea
For the variance term, we can show that
\bea
    \mathrm{Var}\bigg\{\frac{\widehat{m}_2(\bx_{u},\by_i)}{\widetilde{f}(\bx_{u},\by_i)}\bigg\}
    &\to&
    \mathrm{Var}\bigg\{\frac{\widehat{m}_2(\bx_{u},\by_i)}{\mathbb{E}\{\widetilde{f}(\bx_{u},\by_i)\}}\bigg\}
    \nn \\
    &=&
    \frac{\mathbb{E}\{\widehat{m}_2^2(\bx_{u},\by_i)\}}{\mathbb{E}^2\{\widetilde{f}(\bx_{u},\by_i)\}}
    -
    \frac{\mathbb{E}^2\{\widehat{m}_2(\bx_{u},\by_i)\}}{\mathbb{E}^2\{\widetilde{f}(\bx_{u},\by_i)\}} \nn \\
     &=& 
     \frac{\mathbb{E}\{\widehat{m}_2^2(\bx_{u},\by_i)\}}{\mathbb{E}^2\{\widetilde{f}(\bx_{u},\by_i)\}} \nn \\
    \mathbb{E}\{\widehat{m}_2^2(\bx_u, \by_i)\}
    &=&
    \frac{1}{h_1^2h_2^2|\mathcal{N}_{u i}^{\mathrm{rdl}}|}
    \mathbb{E}\bigg\{
        K^{2}\bigg(\frac{\|\bx_{v}-\bx_{u}\|_2}{h_1}, \frac{\|\by_{j}-\by_{i}\|_2}{h_2}\bigg) \epsilon_{v, j}^2
    \bigg\}
   \nn \\
   &=&
   \frac{\sigma^2}{h_1h_2|\mathcal{N}_{u i}^{\mathrm{rdl}}|}
   \iint K^2(a,b)f^*_{ui}(ah_1,bh_2)dadb \nn \\
   &=&  \frac{\sigma^2}{h_1h_2|\mathcal{N}_{u i}^{\mathrm{rdl}}|}
   \iint K^2(a,b)\bigg [f^*_{ui}(0,0)+ah_1^2\frac{\partial f^*_{ui}(0,0)}{\partial a}+bh_2^2\frac{\partial f^*_{ui}(0,0)}{\partial b} \nn \\
   && + o(ah_1+bh_2)\bigg ]dadb \nn \\
   &=& O_p((h_1+h_2)mn\pi^2)
\eea

Then $\frac{\widehat{m}_1(\bx_{u},\by_i)}{\widetilde{f}(\bx_{u},\by_i)} = O_p((h_1+h_2)mn\pi^2)$ when general conditions for the kernel function $K$ are satisfied as discussed above. Next we consider $\mathbb{E}\big\{\frac{\widehat{m}_1(\bx_{u}, \by_i)}{\widetilde{f}(\bx_{u}, \by_i)}\big\}$. We slightly abuse the notation by letting $\br_v = \bx_v-\bx_u$ and $\bs_j = \by_j-\by_i$, and maintain using $r_v = \|\bx_v - \bx_u\|_2$ and $s_j=\|\by_j-\by_i\|_2$. By Slutsky's theorem and Lemma~\ref{lemma3}, we have
\bea
   \mathbb{E}\bigg\{\frac{\widehat{m}_1(\bx_u, \by_i)}{\widetilde{f}(\bx_u,\by_i)}\bigg\}
   &=& 
    \frac{\mathbb{E}\big[\sum_{(v,j)\in\mathcal{N}_{ui}^{\mathrm{rdl}}}
        K\big(\frac{\|\bx_v-\bx_u\|_2}{h_1}, 
            \frac{\|\by_j-\by_i\|_2}{h_2}\big)
        \{g(\bx_{v}, \by_j)-g(\bx_{u}, \by_i)\}\big]}
    {\mathbb{E}\big[\sum_{(v,j)\in\mathcal{N}_{ui}^{\mathrm{rdl}}}
    K\big(\frac{\|\bx_v-\bx_u\|_2}{h_1}, 
            \frac{\|\by_j-\by_i\|_2}{h_2}\big)\big]} \nn \\
   &=& 
    \frac{\mathbb{E}_{\br,\bs}
       \big[\sum_{(v,j)\in\mathcal{N}_{ui}^{\mathrm{rdl}}}
        K\big(\frac{r_v}{h_1}, \frac{s_j}{h_2}\big)
            \{g(\bx_{u}+\br_v, \by_i+\bs_j)-g(\bx_{u}, \by_i)\}\big]}
        {\mathbb{E}_{\br,\bs}\big[\sum_{(v,j)\in\mathcal{N}_{ui}^{\mathrm{rdl}}}
    K\big(\frac{r_v}{h_1}, \frac{s_j}{h_2}\big)\big]}  \nn \\
   &=& 
      \frac{\sum_{(v,j)\in\mathcal{N}_{ui}^{\mathrm{rdl}}}
    \mathbb{E}_{\br,\bs} \big[ K\big(\frac{r_v}{h_1}, \frac{s_j}{h_2}\big)\{g(\bx_{u}+\br_v, \by_i+\bs_j)-g(\bx_{u}, \by_i)\}\big]}
            {\sum_{(v,j)\in\mathcal{N}_{ui}^{\mathrm{rdl}}}
    \mathbb{E}_{r,s}\big\{K\big(\frac{r_v}{h_1}, \frac{s_j}{h_2}\big)\big\}}  \nn \\
   &\leqslant&
   L\cdot\frac{ \sum_{(v,j)\in\mathcal{N}_{ui}^{\mathrm{rdl}}}
    \mathbb{E}_{r,s}\{K \big(\frac{r_v}{h_1}, \frac{s_j}{h_2}\big) (r_v+s_j)\}}
                {\sum_{(v,j)\in\mathcal{N}_{ui}^{\mathrm{rdl}}}
    \mathbb{E}_{r,s}\{K\big(\frac{r_v}{h_1}, \frac{s_j}{h_2}\big)\}} \nn \\
    &=&
    L\cdot\frac{\iint K(a,b)(ah_1+bh_2)f^*_{ui}(ah_1,bh_2)dadb}
                {\iint K(a,b)f^*_{ui}(ah_1, bh_2)dadb} \nn \\
    &\leqslant&
    L(h_1+h_2) \cdot\frac{\iint K(a,b)(a+b)f^*_{ui}(ah_1,bh_2)dadb}
                {\iint K(a,b)f^*_{ui}(ah_1, bh_2)dadb} \nn \\
    &=& O_p(L(h_1+h_2)) \nn \\
    &=& O_p(\sqrt{mk/n}+\sqrt{nk/m})(h_1+h_2) \nn
\eea
when general conditions for the kernel function $K$ are satisfied as discussed above. Similarly, we then consider variance,
\bea
    \mathrm{Var}\bigg\{\frac{\widehat{m}_1(\bx_{u},\by_i)}{\widetilde{f}(\bx_{u},\by_i)}\bigg\} 
    &=& 
    \frac{\mathbb{E}\{\widehat{m}_1^2(\bx_{u},\by_i)\}}{\mathbb{E}^2\{\widetilde{f}(\bx_{u},\by_i)\}} \nn \\
    \mathbb{E} \{\widehat{m}_1^2 (\bx_u, \by_i)\}
    &=&
    \frac{1}{h_1^2h_2^2}
    \mathbb{E}\bigg[
    K\big(\frac{r_v}{h_1}, \frac{s_j}{h_2}\big)
    \{g(\bx_u+\br_v, \by_i+\bs_j)-g(\bx_{u}, \by_i)\} \times \nn \\
    && K\big(\frac{r_{v'}}{h_1}, \frac{s_{j'}}{h_2}\big)
    \{g(\bx_u+\br_{v'}, \by_i+\bs_{j'})-g(\bx_{u}, \by_i)\}
    \bigg] \nn \\
    &\leqslant & \frac{L^2}{h_1^2h_2^2}\mathbb{E}\bigg[K\big(\frac{r_v}{h_1}, \frac{s_j}{h_2}\big)(r_v+s_j) K\big(\frac{r_{v'}}{h_1}, \frac{s_{j'}}{h_2}\big) (r_{v'}+s_{j'})\bigg] \nn \\
    &=& L^2\iiiint K(a,b)K(a',b')(ah_1+bh_2) (a'h_1+b'h_2)dadbda'db' \nn \\
    &\leqslant&
    L^2(h_1+h_2)^2\iiiint K(a,b)K(a',b')(a+b)(a'+b')dadbda'db' \nn \\
    &=& O_p(\sqrt{mk/n}+\sqrt{nk/m})^2(h_1+h_2)^2 \nn
\eea
Based on equation (\ref{expectation}), we prove the convergence of $\widehat{z}_{u,i} \to z_{u,i}$.

\subsection{Proof of Theorem~\ref{theorem2}}
Similar to Lemma~\ref{lemma3}, we can use the same notations to define
\bea
    \dbtilde{f}(\bx_u, \by_i) 
    &=& 
    \frac{1}{h_1h_2|\mathcal{N}_{ui}^{\mathrm{Rdl}}|}
    \sum_{(v,j)\in \mathcal{N}_{ui}^{\mathrm{Rdl}}}
    K\bigg(\frac{\sqrt{\widehat{d}_{u,v}^2 - 2\widehat{\sigma}^2} }{h_1}, \frac{\sqrt{\widehat{d}_{i,j}^2 - 2\widehat{\sigma}^2} }{h_2}\bigg) \nn \\
    &=&
    \frac{1}{h_1h_2|\mathcal{N}_{ui}^{\mathrm{Rdl}}|}
    \sum_{(v,j)\in \mathcal{N}_{ui}^{\mathrm{Rdl}}}
    K\bigg(\frac{\sqrt{\widehat{r}_{v}^2 - 2\widehat{\sigma}^2}}{h_1}, \frac{\sqrt{\widehat{s}_{j}^2 - 2\widehat{\sigma}^2}}{h_2}\bigg) \nn
\eea
and then
\bea
    K\bigg(\frac{\sqrt{\widehat{r}_{v}^2- 2\widehat{\sigma}^2}}{h_1}, \frac{\sqrt{\widehat{s}_{j}^2- 2\widehat{\sigma}^2}}{h_2}\bigg)
    &=& 
    K\bigg(\frac{r_v}{h_1}+\frac{\sqrt{\widehat{r}_v^2-2\widehat{\sigma}^2}-r_v}{h_1}, \frac{s_j}{h_2}+\frac{\sqrt{\widehat{s}_j^2-2\widehat{\sigma}^2}-s_j}{h_2}\bigg) \nn \\
    &=&
    K\bigg(\frac{r_v}{h_1}, \frac{s_j}{h_2}\bigg) + 
    \frac{\partial K\big(\frac{r_v}{h_1}, \frac{s_j}{h_2}\big)}{\partial \frac{r_v}{h_1}}\frac{\sqrt{\widehat{r}_v^2-2\widehat{\sigma}^2}-r_v}{h_1} +
    \frac{\partial K\big(\frac{r_v}{h_1}, \frac{s_j}{h_2}\big)}{\partial \frac{s_j}{h_2}}\frac{\sqrt{\widehat{s}_j^2-2\widehat{\sigma}^2}-s_j}{h_2} \nn
\eea
by Taylor expansion. Then,
\bea
    \dbtilde{f}(\bx_u, \by_i)
    &=&
    \frac{1}{h_1h_2|\mathcal{N}_{ui}^{\mathrm{Rdl}}|}
    \sum_{(v,j)\in \mathcal{N}_{ui}^{\mathrm{Rdl}}}
    K\bigg(\frac{r_v}{h_1}, \frac{s_j}{h_2}\bigg) +
    \frac{1}{h_1h_2|\mathcal{N}_{ui}^{\mathrm{Rdl}}|}
    \sum_{(v,j)\in \mathcal{N}_{ui}^{\mathrm{Rdl}}}
    \frac{\partial K\big(\frac{r_v}{h_1}, \frac{s_j}{h_2}\big)}{\partial \frac{r_v}{h_1}}\frac{\sqrt{\widehat{r}_v^2- 2\widehat{\sigma}^2}-r_v}{h_1} \nn \\
    && +
    \frac{1}{h_1h_2|\mathcal{N}_{ui}^{\mathrm{Rdl}}|}
    \sum_{(v,j)\in \mathcal{N}_{ui}^{\mathrm{Rdl}}}
    \frac{\partial K\big(\frac{r_v}{h_1}, \frac{s_j}{h_2}\big)}{\partial \frac{s_j}{h_2}}\frac{\sqrt{\widehat{s}_j^2- 2\widehat{\sigma}^2}-s_j}{h_2} \nn
\eea

By Propositions~\ref{prop2}, \ref{prop3}, \ref{prop4}, $h_1 = O_p(\sqrt{\widehat{r}_v^2- 2\widehat{\sigma}^2}- r_v)$ and $h_2 = O_p(\sqrt{\widehat{s}_j^2- 2\widehat{\sigma}^2}- s_j)$, and uniform bound for the partial derivation of the kernel function $K$, then
\bea
    \dbtilde{f}(\bx_u, \by_i)
    &\to&
    \frac{1}{h_1h_2|\mathcal{N}_{ui}^{\mathrm{Rdl}}|}
    \sum_{(v,j)\in \mathcal{N}_{ui}^{\mathrm{Rdl}}}
    K\bigg(\frac{r_v}{h_1}, \frac{s_j}{h_2}\bigg) \nn
\eea

Then we can similarly claim that $\frac{\dbtilde{f}(\bx_u, \by_i)}{\mathbb{E}\{\dbtilde{f}(\bx_u, \by_i)\}}\stackrel{p}{\rightarrow} 1$. Recall the estimator defined in equation~(\ref{z_tilde}) and apply Taylor expansion to both its denominator and numerator, we get
\bea
    \widetilde{z}_{u,i}
    &\to&
    \sum_{(v, j) \in \mathcal{N}_{ui}^{\mathrm {Rdl }}}
    \frac{K\big(\frac{r_{v}}{h_1}, 
                \frac{s_{j}}{h_2}\big)}
        {\sum_{(v, j) \in \mathcal{N}_{ui}^{\mathrm {Rdl }}} K\big(\frac{r_{v}}{h_1}, 
                \frac{s_{j}}{h_2}\big) } a_{v,j} \nn \\
    &=&
    \sum_{(v, j) \in \mathcal{N}_{ui}^{\mathrm {Rdl }}}
    \frac{K\big(\frac{r_{v}}{h_1}, 
                \frac{s_{j}}{h_2}\big)}
        {\sum_{(v, j) \in \mathcal{N}_{ui}^{\mathrm {Rdl }}} K\big(\frac{r_{v}}{h_1}, 
                \frac{s_{j}}{h_2}\big) } z_{v,j} 
    + 
    \sum_{(v, j) \in \mathcal{N}_{ui}^{\mathrm {Rdl }}}
    \frac{K\big(\frac{r_{v}}{h_1}, 
                \frac{s_{j}}{h_2}\big)}
        {\sum_{(v, j) \in \mathcal{N}_{ui}^{\mathrm {Rdl }}} K\big(\frac{r_{v}}{h_1}, 
                \frac{s_{j}}{h_2}\big) } \epsilon_{v,j} \nn \\
    &=&
    \widehat{z}_{v,j} + 
    \sum_{(v, j) \in \mathcal{N}_{ui}^{\mathrm {Rdl }}}
    \frac{K\big(\frac{r_{v}}{h_1}, 
                \frac{s_{j}}{h_2}\big)}
        {\sum_{(v, j) \in \mathcal{N}_{ui}^{\mathrm {Rdl }}} K\big(\frac{r_{v}}{h_1}, 
                \frac{s_{j}}{h_2}\big) } \epsilon_{v,j} \nn
\eea
so that we can focus on the second term above which contains $\epsilon_{v,j}$ to provide the convergence. We define it as $\Delta$ for simplicity, then
\bea
    \Delta := \sum_{(v, j) \in \mathcal{N}_{ui}^{\mathrm {Rdl }}}
    \frac{K\big(\frac{r_{u}}{h_1}, 
                \frac{s_{j}}{h_2}\big)}
        {\sum_{(v, j) \in \mathcal{N}_{ui}^{\mathrm {Rdl }}} K\big(\frac{r_{u}}{h_1}, 
                \frac{s_{j}}{h_2}\big) } \epsilon_{v,j}
    \to
    \sum_{(v, j) \in \mathcal{N}_{ui}^{\mathrm {Rdl }}}
    \frac{K\big(\frac{r_{u}}{h_1}, 
                \frac{s_{j}}{h_2}\big)}
        {h_1h_2|\mathcal{N}_{ui}^{\mathrm{Rdl}}| \mathbb{E}\big\{\dbtilde{f}(\bx_u, \by_i)\big\} } \epsilon_{v,j} \nn
\eea

By the assumption that $\epsilon_{v,j}$ is i.i.d and $\mathbb{E}(\epsilon_{v,j})=0$, and the kernel function and $\epsilon_{v,j}$ are mutually independent,
\bea
    \mathbb{E}(\Delta) &=& 0 \nn
\eea
\bea
    \mathrm{Var}(\Delta)
    &=& 
    \frac{ \mathbb{E}^2\big\{K\big(\frac{r_v}{h_1},   
            \frac{s_j}{h_2}\big)\big\}\mathrm{Var}(\epsilon_{v,j}) +
        \mathrm{Var}\big\{K\big(\frac{r_v}{h_1}, \frac{s_j}{h_2}\big)\big\}\mathrm{Var}(\epsilon_{v,j}) +
        \mathrm{Var}\big\{K\big(\frac{r_v}{h_1}, \frac{s_j}{h_2}\big)\big\}\mathbb{E}^2(\epsilon_{v,j}) }
    {h_1^2h_2^2|\mathcal{N}_{ui}^{\mathrm{Rdl}}| \mathbb{E}^2\big\{\dbtilde{f}(\bx_u, \by_i)\big\}} \nn \\
    &=&
    \frac{\mathbb{E}\big\{K^2\big(\frac{r_v}{h_1},   
            \frac{s_j}{h_2}\big)\big\}\cdot \sigma^2 }
    {h_1^2h_2^2|\mathcal{N}_{ui}^{\mathrm{Rdl}}| \mathbb{E}^2\big\{\dbtilde{f}(\bx_u, \by_i)\big\}} \nn \\
    &=&
    \frac{\sigma^2h_1h_2\iint K^2(a, b) f^*_{u,i}(ah_1, bh_2)dadb }
    {h_1^2h_2^2|\mathcal{N}_{ui}^{\mathrm{Rdl}}| \{\iint K(a,b) f^*_{u,i}(ah_1, bh_2)dadb\}^2 } \nn \\
    &=& O_p(h_1h_2mn\pi^2) \nn
\eea
The convergence of $\widetilde{z}_{u,i}\to\widehat{z}_{u,i}$ is hereby obtained.

\end{document}